\documentclass[lettersize,journal]{IEEEtran}
\usepackage{amsmath,amsfonts}
\usepackage{algorithm}
\usepackage{array}
\usepackage[caption=false,font=normalsize,labelfont=sf,textfont=sf]{subfig}
\usepackage{textcomp}
\usepackage{stfloats}
\usepackage{url}
\usepackage{verbatim}
\usepackage{graphicx}
\usepackage{cite}
\usepackage{booktabs}
\usepackage{color}
\usepackage{bbm}
\usepackage{algorithmicx}
\usepackage{algpseudocode}
\usepackage{verbatim}
\usepackage[table]{xcolor}
\usepackage{colortbl}
\usepackage{makecell}
\usepackage{pifont}
\usepackage{bbding}
\usepackage{amsmath}
\usepackage{bm}
\usepackage{multirow}
\usepackage[normalem]{ulem}
\useunder{\uline}{\ul}{}
\usepackage{orcidlink}
\usepackage{hyperref}
\hypersetup{
    colorlinks=true,
    citecolor=blue,
    linkcolor=blue,
    filecolor=magenta,      
    urlcolor=cyan,
}

\usepackage{cleveref} 
\crefname{figure}{Fig.}{Figs.}
\crefname{table}{Tab.}{Tabs.}
\crefname{section}{Sec.}{Secs.}

\usepackage{helvet}
\newcommand{\arial}[1]{{\fontfamily{phv}\selectfont #1}}

\definecolor{bestcolor}{RGB}{255,0,0}      
\definecolor{suboptcolor}{RGB}{9,136,66}   

\begin{document}

\title{0.5\%$>$100\%: Bidirectional Reciprocal Learning for Referring Image Segmentation}

\author{
Xiaoqiang Lu\orcidlink{0000-0003-0338-7645},~\IEEEmembership{Member,~IEEE,}
Licheng Jiao\orcidlink{0000-0003-3354-9617},~\IEEEmembership{Life Fellow,~IEEE,}
Lingling Li\orcidlink{0000-0002-6130-2518},~\IEEEmembership{Senior Member,~IEEE,} \\
Yuting Yang\orcidlink{0000-0002-6720-4134},~\IEEEmembership{Member,~IEEE,}
Long Sun\orcidlink{0000-0001-7446-9336},~\IEEEmembership{Member,~IEEE,} 
Wenping Ma\orcidlink{0000-0001-6569-3029},~\IEEEmembership{Senior Member,~IEEE,} \\
Xu Liu\orcidlink{0000-0002-8780-5455},~\IEEEmembership{Senior Member,~IEEE,}
and Fang Liu\orcidlink{0000-0002-5669-9354},~\IEEEmembership{Senior Member,~IEEE}
\thanks{This work was supported in part by the Joint Funds of the National Natural Science Foundation of China (U22B2054); in part by the National Natural Science Foundation of China (62076192, 62276199, 62431020 and 62276201); in part by the 111 Project; in part by the Program for Cheung Kong Scholars and Innovative Research Team in University (IRT 15R53); in part by the National Science Basic Research Plan in Shaanxi Province of China under Grant 2019JQ-659 and 2022JQ-607; in part by the Science and Technology Innovation Project from the Chinese Ministry of Education; in part by the National Key Laboratory of Human-Machine Hybrid Augmented Intelligence, Xi'an Jiaotong University (HMHAI-202404 and HMHAI-202405). \textit{(Corresponding author: Licheng Jiao.)}}

\thanks{The authors are with the Key Laboratory of Intelligent Perception and Image Understanding of Ministry of Education, International Research Center for Intelligent Perception and Computation, School of Artificial Intelligence, Xidian University, Xi’an, Shaanxi Province, 710071, China (e-mail: luxiaoqiang5903@163.com; lchjiao@mail.xidian.edu.cn).}
}

\markboth{Submitted to IEEE}%
{Shell \MakeLowercase{\textit{et al.}}: A Sample Article Using IEEEtran.cls for IEEE Journals}

\IEEEpubid{0000--0000/00\$00.00~\copyright~2026 IEEE}

\maketitle

\begin{abstract}
    Recent advances in vision foundation models (VFMs) have shown remarkable capabilities across diverse unimodal visual tasks. However, adapting VFMs to referring image segmentation (RIS) typically necessitates precise vision-language alignment via full fine-tuning, incurring substantial computational overhead and risking catastrophic forgetting. While existing parameter-efficient fine-tuning (PEFT) methods enable safe knowledge transfer with minimal training costs, they predominantly operate independently within individual modalities or focus exclusively on unidirectional guidance from language to vision, overlooking progressive cross-modal interaction and visual feedback for textual refinement. To address these limitations, we propose \textbf{B}idirectional \textbf{R}eciprocal \textbf{L}earning (\textbf{BRL}), a novel adapter-based PEFT framework that facilitates hierarchical, bidirectional information flow within both token-mixing and channel-mixing layers of frozen foundation models. Specifically, BRL introduces two complementary lightweight modules. The Reciprocal Attention Adapter (RAA) performs cross-modal query-key exchanges at the token level, enabling visual and linguistic tokens to mutually attend to each other for fine-grained spatial grounding. The Reciprocal Gate Adapter (RGA) generates cross-modal gating signals at the channel level, allowing global semantic context from one modality to adaptively recalibrate channel activations of the other. Extensive experiments on RefCOCO, RefCOCO+, and RefCOCOg benchmarks demonstrate the superiority of BRL over prior RIS methods, achieving state-of-the-art performance while requiring less than 0.5\% backbone parameter updates. Code and models will be released at \url{https://github.com/xiaoqiang-lu/BRL}.
\end{abstract}


\section{Introduction}
\label{introduction}

\IEEEPARstart{R}{eferring} image segmentation (RIS) aims to segment the specific visual object corresponding to a free-form textual expression~\cite{lavt_tpami, efn_tpami, vlt}. As a foundational dense prediction task bridging vision and language, RIS brings crucial effects in human-computer interaction~\cite{soc, lisa, segllm, popen}, assisted diagnosis~\cite{lvit, reclmis, tvenet}, and disaster assessment~\cite{rrsecs, seeformer, lscf}. In contrast to traditional image segmentation~\cite{mask2former, lsst, upernet, speed}, which learns mappings to a set of semantic categories~\cite{deeplabv3+, wscl, segformer, umcl}, RIS requires establishing precise contextual relationships between visual and linguistic cues. To achieve it, pioneering works follow a two-stage paradigm that identifies the most relevant proposal according to its similarity with the expression~\cite{cmatt, cmpc_tpami}, whereas later studies have shifted toward a one-stage paradigm that enables end-to-end comprehension via multimodal feature fusion and interaction~\cite{lavt, polyformer, magnet, remamber, lqmformer}.

\begin{figure}[t]
    \centering
    \includegraphics[width=1\linewidth]{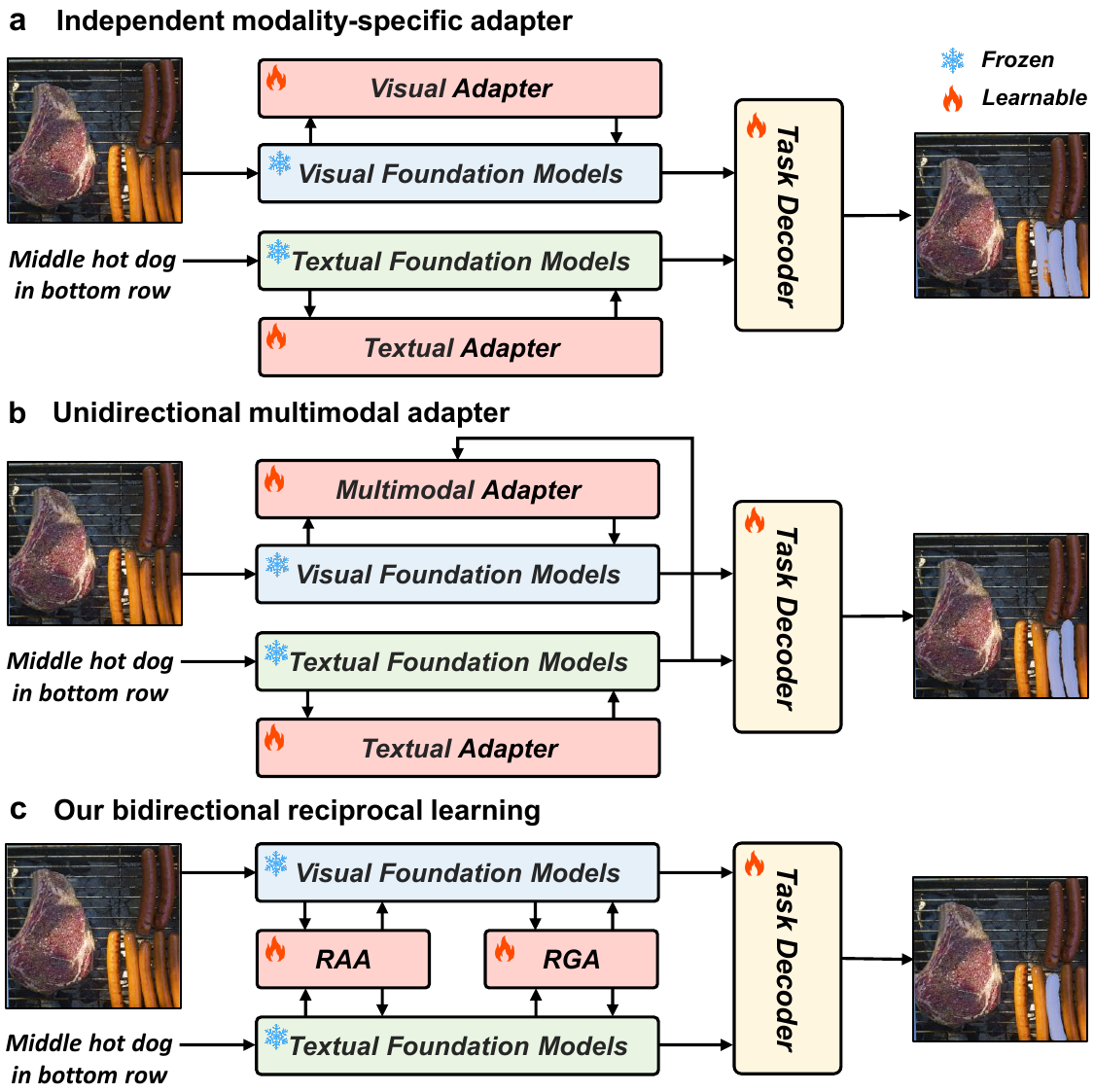}
    \caption{Comparisons of different PEFT methods for RIS. \arial{\textbf{a}}: Independent modality-specific adapters operate separately on visual and textual streams without cross-modal interaction. \arial{\textbf{b}}: Unidirectional multimodal adapter enables limited fusion due to its only supporting guidance from language to vision. \arial{\textbf{c}}: Unlike previous methods that treat language as static guidance, our BRL formulates multimodal adaptation as a reciprocal representation learning process via the reciprocal attention adapter and reciprocal gate adapter.}
    \label{fig_paradigm}
\end{figure}

\IEEEpubidadjcol 

Recent research has increasingly leveraged multimodal large language models (MLLMs)~\cite{clip, llava} to advance RIS by exploiting their pre-aligned semantic spaces for cross-modal reasoning~\cite{etris, barleria, risclip, lisa, segllm, popen}. Nevertheless, prevailing MLLMs are mainly pretrained via global image-text matching, yielding coarse-grained representations suboptimal for RIS, which demands fine-grained perception and precise localization. In contrast, vision foundation models (VFMs)~\cite{sam, dinov2, dinov3} have achieved remarkable success across diverse visual tasks, driven by their superior capacity for capturing fine-grained structural details. However, adapting these powerful unimodal VFMs to the multimodal RIS scenario presents significant obstacles. Traditional full fine-tuning incurs prohibitive computational overhead, while the discrepancy between unimodal pre-training objectives and multimodal feature alignment often risks catastrophic forgetting and representation interference.

\begin{figure*}[t]
    \centering
    \includegraphics[width=1\linewidth]{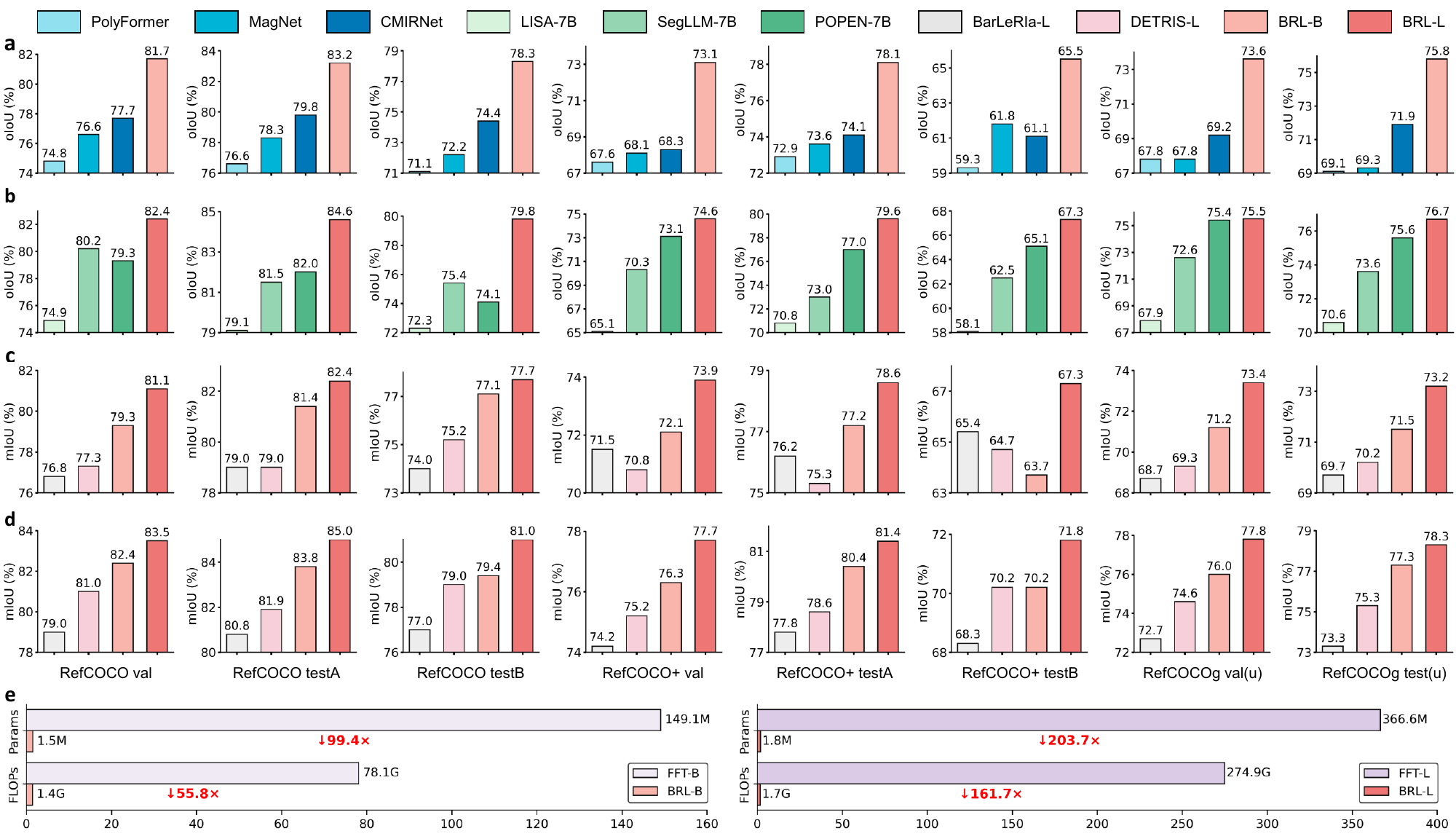}
    \caption{Comparisons of BRL with state-of-the-art methods across RefCOCO, RefCOCO+, and RefCOCOg datasets. \arial{\textbf{a}}: Compared with traditional full fine-tuning models using the mixed RefCOCO/+/g setting, our BRL with DINOv3-B~\cite{dinov3} (BRL-B) achieves substantial performance gains of oIoU across all RIS benchmarks. \arial{\textbf{b}}: Compared with multimodal large language models using the mixed setting, our BRL with DINOv3-L (BRL-L) outperforms them across all benchmarks while updating merely 1.8M parameters within encoders. \arial{\textbf{c}}: Compared with parameter-efficient fine-tuning (PEFT) methods using the common setting, BRL-B surpasses DETRIS-L~\cite{detris} on seven benchmarks except RefCOCO+ testB. \arial{\textbf{d}}: Compared with PEFT methods using the mixed setting, BRL-L brings consistent performance gains of mIoU across all RIS benchmarks. \arial{\textbf{e}}: Compared with full fine-tuning (FFT) foundation models, BRL reduces both learnable parameters and FLOPs by nearly 100× and 200× under the base and large architectures, respectively.}
    \label{fig_comp_bar}
\end{figure*}

In response, parameter-efficient fine-tuning (PEFT) has emerged as a promising alternative, seeking to safely condense general knowledge into task-specific expertise while minimizing training costs~\cite{peft_llm, peft_vfm, peft_fm}. Current PEFT methods mainly include reparameterization-based~\cite{lora, lorand, qlora}, prompt-based~\cite{vpt, qformer}, and adapter-based~\cite{vitadapter, clipadapter} tuning. Nevertheless, most of them exhibit critical issues when applied to RIS. As illustrated in \cref{fig_paradigm}.\arial{\textbf{a}}, common adapter-based methods~\cite{adapter, adaptformer} operate independently within individual modalities, lacking explicit cross-modal interaction. Recent work such as DETRIS~\cite{detris} explores the feasibility of VFMs in RIS by constructing a multimodal adapter using global textual information, as depicted in \cref{fig_paradigm}.\arial{\textbf{b}}. While effective, it remains constrained by two fundamental limitations. First, the language representation remains static during multimodal interaction, preventing ambiguous expressions from being grounded and refined according to visual evidence. Second, the absence of inter-modality progressive learning may lead to the accumulation of negative effects caused by misaligned representations.

Considering the current progress and the above two shortcomings, a natural question arises: \emph{can we unleash the full potential of VFMs in multimodal dense predictions?} We answer it affirmatively with our proposed Bidirectional Reciprocal Learning (BRL) framework, a lightweight yet efficient adapter-based tuning method that enables vision-language interaction bidirectionally and hierarchically. As shown in \cref{fig_paradigm}.\arial{\textbf{c}}, BRL consists of two core components: the Reciprocal Attention Adapter (RAA), which facilitates token-wise mutual reasoning for dense cross-modal alignment, and the Reciprocal Gate Adapter (RGA), which realizes channel-wise reciprocal regulation to balance modality contributions and suppress irrelevant signals. Specifically, RAA introduces a reciprocal attention mechanism in the self-attention layer, which leverages linguistic queries to guide the localization of visual tokens while simultaneously feeding visual evidence back to refine textual semantics. This mutual refinement builds second-order dependencies between modalities, allowing visual grounding and linguistic semantics to evolve jointly rather than relying on static language priors. In contrast, RGA operates on the feed-forward layer by establishing a reciprocal gating mechanism that dynamically adjusts activation strengths based on cross-modal global prior information. This gated regulation mitigates cross-modal representation imbalance, attenuates modality-specific redundancy, and amplifies informative responses, leading to more robust multimodal representations. Notably, RAA and RGA are complementary: the former focuses on spatial correspondence learning, whereas the latter regulates global semantic consistency and feature selectivity.

To evaluate the effectiveness of our method, we conduct extensive experiments on RefCOCO, RefCOCO+, and RefCOCOg benchmarks. As summarized in \cref{fig_comp_bar}, BRL not only surpasses traditional full fine-tuning methods and recent MLLM-based methods, but also dramatically improves adaptation efficiency by reducing learnable parameters and FLOPs by nearly 100$\times$ and 200$\times$ under the base and large architectures, respectively. These results demonstrate the remarkable potential of reciprocal multimodal adaptation for efficient dense vision-language understanding. Our main contributions are as follows:

\begin{enumerate}
    \item We propose BRL, a parameter-efficient learning framework that unleashes the power of vision foundation models in referring image segmentation, allowing progressive mutual vision-language alignment and refinement while updating less than 0.5\% parameters.

    \item We introduce RAA, a lightweight token-level interaction module that performs reciprocal cross-modal attention within transformer self-attention layers, enabling visual and linguistic tokens to mutually refine each other for fine-grained spatial grounding.

    \item We further develop RGA, a channel-level modulation module that dynamically generates cross-modal gating signals to adaptively recalibrate heterogeneous feature responses, thereby suppressing modality-specific redundancy and enhancing semantic consistency.
\end{enumerate}

\section{Related Work}
\label{related_work}

\subsection{Referring Image Segmentation}

Referring image segmentation (RIS) aims to localize and segment the target region in an image according to a free-form natural language expression~\cite{refcoco, gref1, gref2}. As a representative multimodal dense prediction task, RIS requires simultaneously understanding visual structures and linguistic semantics while establishing fine-grained cross-modal correspondence between them. Early RIS methods mainly follow a two-stage paradigm, where candidate object proposals are first generated and then matched with the referring expression based on multimodal similarity~\cite{mattnet, cmatt}. Although effective, such approaches heavily depend on proposal quality and often suffer from limited end-to-end optimization capability.

To overcome these limitations, recent studies gradually transition toward one-stage end-to-end frameworks~\cite{vlt, efn, restr}, which directly perform multimodal interaction and dense prediction within unified architectures. Benefiting from transformer-based multimodal reasoning, these methods fuse visual and linguistic tokens through cross-attention mechanisms~\cite{rela, magnet, lqmformer}, significantly improving global contextual understanding and spatial grounding ability. More recently, the emergence of multimodal large language models (MLLMs)~\cite{clip, llava} has further promoted RIS by leveraging large-scale vision-language pretraining and aligned semantic representations~\cite{lisa, gsva, segllm, popen}. Despite their impressive performance, most existing MLLM-based methods rely on careful fine-tuning or heavy multimodal instruction tuning, resulting in substantial computational overhead and potential risks of catastrophic forgetting. Moreover, prevailing multimodal pretraining objectives mainly emphasize global image-text alignment, which remains suboptimal for RIS requiring fine-grained spatial localization and dense cross-modal interaction.

In contrast, recent vision foundation models (VFMs)~\cite{sam, dinov2, dinov3} demonstrate remarkable capability in capturing fine-grained visual structures and dense spatial details, making them highly promising for RIS. However, effectively adapting powerful unimodal VFMs to multimodal dense prediction remains insufficiently explored, especially under parameter-efficient learning paradigms.

\subsection{Parameter-Efficient Fine-Tuning}

Parameter-efficient fine-tuning (PEFT) aims to adapt large pre-trained models to downstream tasks by updating only a small subset of parameters while preserving the majority of pretrained knowledge~\cite{peft_fm}. Existing PEFT methods mainly include adapter-based~\cite{adapter, adaptformer, clipadapter}, prompt-based~\cite{prompt_tuning, vpt, qformer}, and reparameterization-based learning~\cite{lora, lora_fa, lorand}. Adapter-based methods insert lightweight bottleneck modules into transformer layers for task-specific adaptation~\cite{adapter}, whereas prompt-based methods optimize learnable prompts to guide downstream learning~\cite{prompt_tuning}. Reparameterization-based approaches instead approximate weight updates through low-rank decomposition for efficient optimization~\cite{lora}. Owing to their strong efficiency and transferability, PEFT techniques have achieved remarkable success in large language models~\cite{peft_llm} and vision foundation models~\cite{peft_vfm}.

Nevertheless, applying PEFT to multimodal dense prediction remains highly challenging. Unlike conventional classification or generation tasks, RIS requires fine-grained reciprocal interaction between visual grounding and linguistic semantics throughout the adaptation process. Existing RIS-oriented PEFT methods mainly focus on modality-specific adaptation, such as inserting adapters only into visual encoders or employing lightweight language modules for textual refinement. A few recent works~\cite{detris, risclip} further explore multimodal adaptation for RIS by injecting linguistic guidance into visual representations. However, their interactions are typically shallow or unidirectional, where language mainly serves as static semantic supervision for visual modulation. Such asymmetric adaptation prevents multimodal representations from being progressively refined through reciprocal feedback and limits the establishment of hierarchical cross-modal dependencies.

Different from existing PEFT paradigms, our proposed BRL formulates multimodal adaptation as a reciprocal representation learning process. By jointly modeling token-level reciprocal reasoning and channel-level reciprocal regulation, BRL enables progressive bidirectional interaction between frozen vision and language foundation models for efficient and robust referring image segmentation.

\section{Method}
\label{method}

\begin{figure*}[t]
    \centering
    \includegraphics[width=1\linewidth]{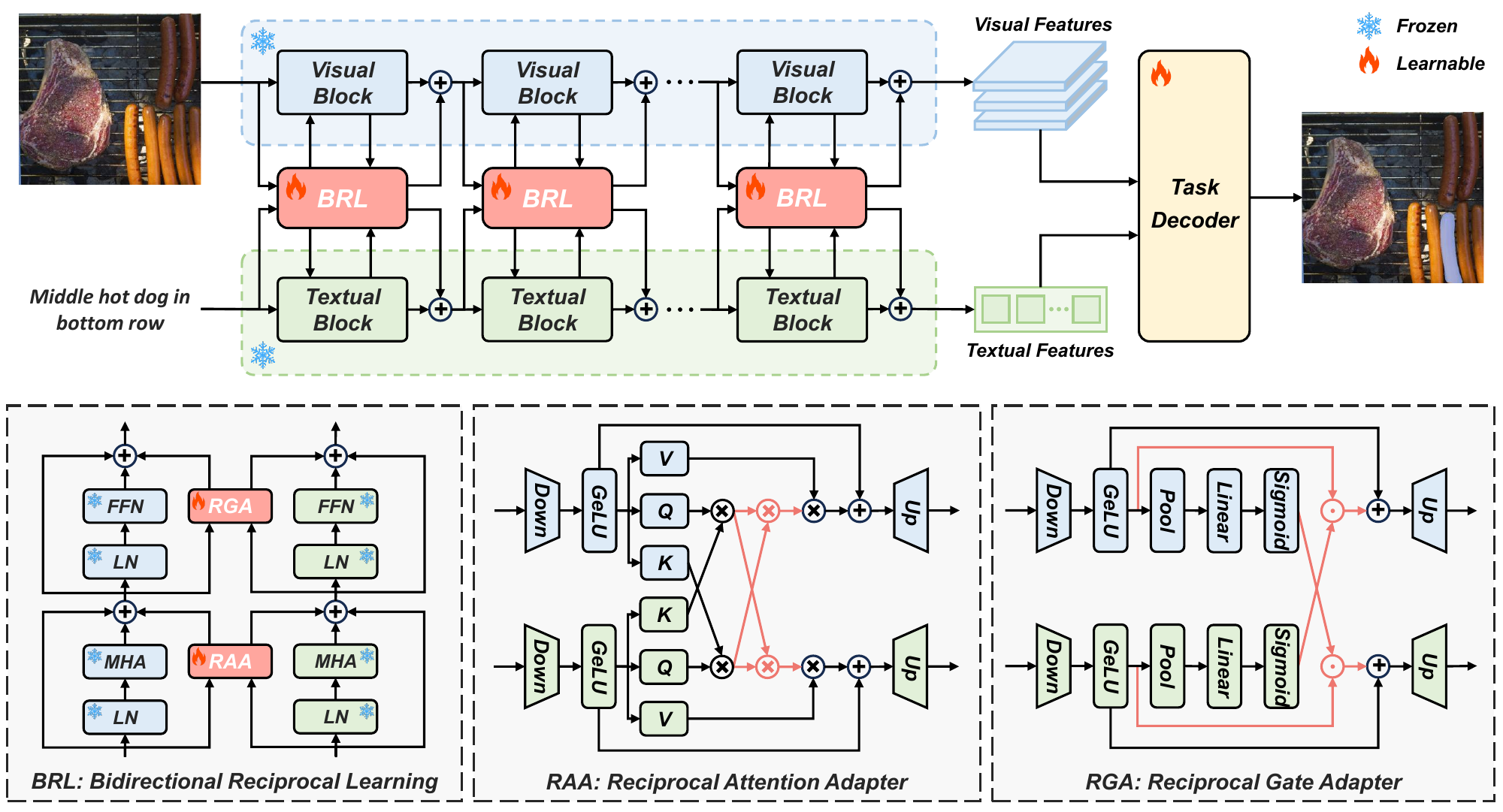}
    \caption{Overview of the proposed Bidirectional Reciprocal Learning (BRL) framework. BRL inserts two lightweight reciprocal adapters, namely the Reciprocal Attention Adapter (RAA) and the Reciprocal Gate Adapter (RGA), into frozen vision and language foundation models for efficient multimodal adaptation. Specifically, RAA performs token-level reciprocal interaction via bidirectional attention composition to establish fine-grained cross-modal correspondence, while RGA conducts channel-level reciprocal regulation through cross-modal gated modulation to enhance global semantic consistency and adaptive feature selectivity. Through progressive bidirectional adaptation across selected transformer layers, BRL enables hierarchical reciprocal representation learning between visual grounding and linguistic semantics with minimal trainable parameters.}
    \label{fig_brl}
\end{figure*}

\subsection{Bidirectional Reciprocal Learning Overview}

Given an image $I$ and a referring expression $T$, the goal of referring image segmentation (RIS) is to predict the target binary mask $M \in \mathbb{R}^{H \times W}$ corresponding to the referred object. Unlike existing parameter-efficient fine-tuning (PEFT) methods that mainly perform unidirectional language-guided visual adaptation, we formulate multimodal adaptation as a \emph{reciprocal representation learning} process, where visual and linguistic representations are progressively refined through bidirectional interaction.

As illustrated in Fig.~\ref{fig_brl}, the proposed \textbf{Bidirectional Reciprocal Learning (BRL)} framework consists of two frozen foundation backbones: a vision foundation model (VFM) $\mathcal{V}$ and a language foundation model (LFM) $\mathcal{L}$. Given the visual input $I$, the visual encoder extracts hierarchical visual tokens:

\begin{equation}
    \mathbf{V} = \mathcal{V}(I), 
    \quad
    \mathbf{V} \in \mathbb{R}^{N_v \times C_v},
\end{equation}
where $N_v$ and $C_v$ denote the token number and feature dimension, respectively. Similarly, the language encoder transforms the referring expression $T$ into linguistic representations:

\begin{equation}
    \mathbf{L} = \mathcal{L}(T), 
    \quad
    \mathbf{L} \in \mathbb{R}^{N_l \times C_l},
\end{equation}
where $N_l$ and $C_l$ represent the number and dimension of linguistic tokens.

Instead of updating the VFM and LFM parameters, BRL inserts lightweight reciprocal adapters into selected transformer layers to achieve efficient multimodal adaptation with minimal trainable computational cost. Specifically, BRL introduces two complementary components: the Reciprocal Attention Adapter (RAA) and the Reciprocal Gate Adapter (RGA). RAA is embedded into transformer self-attention layers to establish token-level reciprocal interaction, whereas RGA is integrated into feed-forward layers to perform channel-level reciprocal regulation. Through hierarchical bidirectional adaptation, BRL progressively constructs reciprocal multimodal dependencies between visual grounding and linguistic semantics. Notably, RAA and RGA operate cooperatively rather than independently: RAA focuses on reciprocal spatial correspondence learning, whereas RGA further stabilizes multimodal adaptation through global semantic regulation.

Formally, given the selected $i$-th transformer layer, BRL first conducts reciprocal token interaction via the RAA:

\begin{equation}
    \mathbf{V}_i^{raa},~\mathbf{L}_i^{raa} = \operatorname{RAA}_i(\mathbf{V}_{i-1},~\mathbf{L}_{i-1}),
\end{equation}
where $\mathbf{V}_{i-1}$ and $\mathbf{L}_{i-1}$ denote the visual and linguistic representations from the previous transformer layer. The reciprocal attention outputs are then integrated into the transformer self-attention residual branches:

\begin{equation}
    \mathbf{V}_i^{att} = \mathbf{V}_{i-1} + \operatorname{MHA}\big(\operatorname{LN}(\mathbf{V}_{i-1})\big) + \alpha \cdot \mathbf{V}_i^{raa},
\end{equation}

\begin{equation}
    \mathbf{L}_i^{att} = \mathbf{L}_{i-1} + \operatorname{MHA}\big(\operatorname{LN}(\mathbf{L}_{i-1})\big) + \alpha \cdot \mathbf{L}_i^{raa},
\end{equation}
where $\operatorname{LN}(\cdot)$ and $\operatorname{MHA}(\cdot)$ denote layer normalization and multi-head self-attention, respectively, and $\alpha$ is a residual scaling factor controlling the reciprocal adaptation strength. Subsequently, BRL further performs reciprocal channel regulation through the RGA:

\begin{equation}
    \mathbf{V}_i^{rga},~\mathbf{L}_i^{rga} = \operatorname{RGA}_i(\mathbf{V}_i^{att},~\mathbf{L}_i^{att}),
\end{equation}
whose outputs are injected into feed-forward layers as:

\begin{equation}
    \mathbf{V}_i = \mathbf{V}_i^{att} + \operatorname{FFN}\big(\operatorname{LN}(\mathbf{V}_i^{att})\big) + \alpha \cdot \mathbf{V}_i^{rga},
\end{equation}

\begin{equation}
    \mathbf{L}_i = \mathbf{L}_i^{att} + \operatorname{FFN}\big(\operatorname{LN}(\mathbf{L}_i^{att})\big) + \alpha \cdot \mathbf{L}_i^{rga},
\end{equation}
where $\operatorname{FFN}(\cdot)$ denotes the feed-forward network. 

Through iterative reciprocal adaptation between RAA and RGA across multiple transformer layers, visual and linguistic representations gradually evolve from independent unimodal features toward semantically aligned multimodal representations, thereby facilitating robust dense cross-modal grounding and understanding. Overall, BRL establishes a hierarchical reciprocal adaptation paradigm, where token-level interaction captures fine-grained spatial correspondence, while channel-level regulation enhances global semantic consistency and adaptive feature selectivity.

\subsection{Reciprocal Token Interaction}

Existing multimodal PEFT methods mainly adopt unidirectional interaction paradigms, where linguistic representations are treated as static semantic guidance for visual modulation. Although such strategies introduce cross-modal information into visual adaptation, the interaction process remains inherently asymmetric, preventing visual grounding and linguistic semantics from being jointly optimized through reciprocal feedback. Moreover, conventional cross-attention mechanisms mainly establish first-order correspondence between modalities, which is insufficient for RIS requiring fine-grained spatial reasoning and semantic disambiguation.

To address these limitations, we propose the \textbf{Reciprocal Attention Adapter (RAA)} to perform bidirectional token-level reciprocal interaction within transformer self-attention layers. Different from conventional cross-modal attention that directly aggregates heterogeneous features, RAA explicitly constructs reciprocal second-order dependencies between visual and linguistic representations, enabling each modality to progressively refine itself through the contextual structure of the other modality.

Given visual and linguistic representations $\mathbf{V}$ and $\mathbf{L}$, RAA first projects them into a shared low-dimensional latent space with channel dimension $d$:

\begin{equation}
    \hat{\mathbf{V}} = \operatorname{GELU}\big(\phi_v(\mathbf{V})\big),
    \quad
    \hat{\mathbf{L}} = \operatorname{GELU}\big(\phi_l(\mathbf{L})\big),
\end{equation}
where
\begin{equation}
    \hat{\mathbf{V}} \in \mathbb{R}^{N_v \times d},
    \quad
    \hat{\mathbf{L}} \in \mathbb{R}^{N_l \times d},
\end{equation}
and $\phi_v(\cdot)$ and $\phi_l(\cdot)$ denote lightweight down-projection layers, $\operatorname{GELU}(\cdot)$ is the GELU activation function. The bottleneck dimension $d$ controls the adaptation capacity and computational complexity of reciprocal interaction.

Based on the projected representations, RAA then constructs bidirectional cross-modal attention maps:

\begin{equation}
    \mathbf{A}_{v \rightarrow l} = \operatorname{Softmax}\left(\frac{\mathbf{Q}_v \mathbf{K}_l^\top}{\sqrt{d}}\right),
\end{equation}

\begin{equation}
    \mathbf{A}_{l \rightarrow v} = \operatorname{Softmax}\left(\frac{\mathbf{Q}_l \mathbf{K}_v^\top}{\sqrt{d}}\right),
\end{equation}
where
\begin{equation}
    \mathbf{Q}_v = \hat{\mathbf{V}}\mathbf{W}_v^Q,
    \quad
    \mathbf{K}_v = \hat{\mathbf{V}}\mathbf{W}_v^K,
\end{equation}

\begin{equation}
    \mathbf{Q}_l = \hat{\mathbf{L}}\mathbf{W}_l^Q,
    \quad
    \mathbf{K}_l = \hat{\mathbf{L}}\mathbf{W}_l^K,
\end{equation}
and $\mathbf{W}_{v/l}^Q$ and $\mathbf{W}_{v/l}^K$ represent learnable query and key projections, respectively. Unlike standard cross-attention mechanisms that directly utilize $\mathbf{A}_{v \rightarrow l}$ or $\mathbf{A}_{l \rightarrow v}$ for feature aggregation, RAA further establishes reciprocal second-order dependencies through attention composition:

\begin{equation}
    \mathbf{A}_v = \mathbf{A}_{v \rightarrow l} \mathbf{A}_{l \rightarrow v},
\end{equation}

\begin{equation}
    \mathbf{A}_l = \mathbf{A}_{l \rightarrow v} \mathbf{A}_{v \rightarrow l}.
\end{equation}

Such reciprocal attention composition enables each modality to establish higher-order contextual dependencies through the opposite modality. Consequently, visual representations can be refined according to linguistic semantic structures, while linguistic representations simultaneously benefit from visual evidence feedback. This reciprocal reasoning mechanism facilitates more robust multimodal alignment and alleviates the representation inconsistency caused by unidirectional adaptation.

Next, the reciprocal attention maps are then utilized for feature aggregation:

\begin{equation}
    \mathbf{V}^{\prime} = \mathbf{A}_v \mathbf{W}_v^V,
    \quad
    \mathbf{L}^{\prime} = \mathbf{A}_l \mathbf{W}_l^V,
\end{equation}
where $\mathbf{W}_{v/l}^V$ are learnable value projections.

Finally, reciprocal token adaptation is performed as:

\begin{equation}
    \mathbf{V}^{raa} = \psi_v(\hat{\mathbf{V}} + \mathbf{V}^{\prime}),
    \quad
    \mathbf{L}^{raa} = \psi_l(\hat{\mathbf{L}} + \mathbf{L}^{\prime}),
\end{equation}
where $\psi_v(\cdot)$ and $\psi_l(\cdot)$ denote lightweight up-projection layers that restore modality-specific feature dimensions.

Compared with typical cross-attention mechanisms, RAA establishes reciprocal token-level reasoning between modalities, enabling fine-grained spatial grounding and semantic disambiguation through progressive bidirectional interaction.

\subsection{Reciprocal Channel Regulation}

Although token-level reciprocal interaction enables fine-grained spatial correspondence learning, multimodal representations in RIS still suffer from modality imbalance and semantic redundancy. In particular, token-wise interaction mainly focuses on local correspondence reasoning, while lacking explicit regulation over global semantic consistency and feature selectivity across channels. Consequently, irrelevant activations or dominant modality biases may progressively accumulate during deep multimodal adaptation.

To address this issue, we propose the \textbf{Reciprocal Gate Adapter (RGA)} to perform bidirectional channel-level reciprocal regulation within transformer feed-forward layers. Different from existing gating mechanisms that independently recalibrate features within a single modality, RGA dynamically generates cross-modal gating priors from global semantic statistics, allowing each modality to adaptively regulate the channel responses of the other modality.

Given visual and linguistic representations after self-attention with reciprocal token interaction, namely $\mathbf{V}^{att}$ and $\mathbf{L}^{att}$, RGA first projects them into a shared low-dimensional latent space:

\begin{equation}
    \hat{\mathbf{V}}^{att} = \operatorname{GELU}\big(\phi_v(\mathbf{V}^{att})\big),~
    \hat{\mathbf{L}}^{att} = \operatorname{GELU}\big(\phi_l(\mathbf{L}^{att})\big),
\end{equation}
where
\begin{equation}
    \hat{\mathbf{V}}^{att} \in \mathbb{R}^{N_v \times d},
    \quad
    \hat{\mathbf{L}}^{att} \in \mathbb{R}^{N_l \times d}.
\end{equation}

To capture global semantic context, RGA then aggregates modality-wise global representations through spatial average pooling:

\begin{equation}
    \mathbf{g}_v = \frac{1}{N_v} \sum_{n=1}^{N_v} \hat{\mathbf{V}}^{att}_{n},
    \quad
    \mathbf{g}_l = \frac{1}{N_l} \sum_{n=1}^{N_l} \hat{\mathbf{L}}^{att}_{n},
\end{equation}
where
\begin{equation}
    \mathbf{g}_v \in \mathbb{R}^{1 \times d},
    \quad
    \mathbf{g}_l \in \mathbb{R}^{1 \times d}.
\end{equation}

Unlike common channel attention mechanisms that generate intra-modal gates independently, RGA performs reciprocal cross-modal gating:

\begin{equation}
    \mathbf{w}_v = \sigma(\mathbf{W}_v \mathbf{g}_v),
    \quad
    \mathbf{w}_l = \sigma(\mathbf{W}_l \mathbf{g}_l),
\end{equation}
where $\sigma(\cdot)$ denotes the sigmoid activation function, and $\mathbf{W}_v$ and $\mathbf{W}_l$ are learnable linear projections. The generated gating weights are then reciprocally injected into the opposite modality:

\begin{equation}
    \mathbf{V}^{\ast} = \hat{\mathbf{V}}^{att} \odot \mathbf{w}_l,
    \quad
    \mathbf{L}^{\ast} = \hat{\mathbf{L}}^{att} \odot \mathbf{w}_v,
\end{equation}
where $\odot$ denotes channel-wise multiplication. Through such reciprocal gating, global semantic priors from one modality dynamically regulate the activation distribution of the other modality, thereby suppressing modality-specific redundancy and enhancing semantically informative responses.

Finally, reciprocal channel adaptation is performed through:
\begin{equation}
    \mathbf{V}^{rga} = \psi_v(\hat{\mathbf{V}}^{att} + \mathbf{V}^{\ast}),
    \quad
    \mathbf{L}^{rga} = \psi_l(\hat{\mathbf{L}}^{att} + \mathbf{L}^{\ast}).
\end{equation}

Compared with conventional gating mechanisms, RGA establishes reciprocal channel-level semantic regulation between modalities, enabling adaptive feature selection and globally consistent multimodal representation learning under lightweight parameter adaptation.

\subsection{Task Decoder and Training Objective}

Following previous RIS-oriented PEFT methods~\cite{cris, etris, detris}, we adopt the same transformer-based decoder to generate the final segmentation mask, ensuring fair comparisons with pioneer works. 

Specifically, the decoder takes multiple visual features $\mathbf{V_i},~i\in\{5,8,11\}$ from three selected layers of the VFM, together with linguistic token representations $\mathbf{L}$ and a global textual state feature $\mathbf{s}$. To establish coarse multimodal correspondence, the decoder first performs cross-modal hierarchical aggregation between the global textual state and multi-stage visual features:

\begin{equation}
    \bar{\mathbf{V}} = \operatorname{Conv}\Big(\operatorname{Concat}\big([\operatorname{CA}(\mathbf{V}_i,~\mathbf{s})]\big)\Big),
\end{equation}
where $\operatorname{CA}(\cdot)$, $\operatorname{Concat}(\cdot)$, and $\operatorname{Conv}(\cdot)$ denote cross-attention, concatenation, and convolution operations, respectively.

The aggregated visual-semantic feature $\bar{\mathbf{V}}$ is subsequently refined through three transformer decoding layers consisting of self-attention, cross-attention, and feed-forward networks:

\begin{equation}
\mathbf{F} = \operatorname{FFN}\Big(\operatorname{CA}\big(\operatorname{SA}(\bar{\mathbf{V}}),~\mathbf{L}\big)\Big),
\end{equation}
where $\operatorname{SA}(\cdot)$ represents self-attention.

Finally, the refined multimodal feature $\mathbf{F}$ is dynamically projected into the segmentation space using text-conditioned convolution kernels generated from the global textual state:

\begin{equation}
    M = \operatorname{Conv}(\mathbf{F}, \Phi(\mathbf{s})),
\end{equation}
where $\Phi(\cdot)$ denotes the dynamic kernel generation function conditioned on the textual state feature.

For training, previous RIS methods commonly introduce an additional text-to-visual contrastive loss to strengthen vision-language alignment~\cite{cris, detris}. In contrast, the proposed BRL already establishes effective reciprocal multimodal alignment within frozen foundation models through bidirectional token interaction and channel modulation. Therefore, we optimize the model using only standard segmentation objectives:

\begin{equation}
    \mathcal{L} = \lambda_1\mathcal{L}_{bce} + \lambda_2 \mathcal{L}_{dice},
\end{equation}
where $\mathcal{L}_{bce}$ and $\mathcal{L}_{dice}$ denote the binary cross-entropy loss and Dice loss, respectively, while $\lambda_1 = \lambda_2 = 1$ are balancing coefficients. Such a simple objective further demonstrates that the performance gains of BRL mainly originate from effective reciprocal multimodal adaptation rather than auxiliary alignment supervision.

\section{Experiments}
\label{experiments}

\subsection{Datasets and Evaluation Metrics}

\subsubsection{Datasets}

We conduct experiments on three widely used referring image segmentation benchmarks:

RefCOCO~\cite{refcoco}: The RefCOCO is a large-scale and widely adopted benchmark for RIS. It contains 19,994 images accompanied by 142,210 referring expressions describing 50,000 objects. All annotations are collected through an interactive two-player game built on the MSCOCO~\cite{mscoco} dataset. The benchmark is split into four partitions: 120,624 samples for training, 10,834 for val, 5,657 for testA, and 5,095 for testB. The expressions are relatively concise, averaging 3.6 words.

RefCOCO+~\cite{refcoco}: The RefCOCO+ provides 141,564 referring expressions related to 49,856 objects in 19,992 images. Unlike RefCOCO, it introduces a higher level of difficulty by deliberately removing absolute spatial terms, forcing models to rely more on appearance cues rather than positional hints. Its data split follows the same protocol as RefCOCO, comprising 120,624 training examples, 10,758 in val, 5,726 in testA, and 4,889 in testB.

RefCOCOg~\cite{gref1, gref2}: The RefCOCOg comprises 104,560 referring expressions that describe 54,822 objects within 26,711 images. The text annotations are relatively long, with an average length of 8.4 words, providing richer information about object appearances and locations. We follow the commonly used Google~\cite{gref1} split and UMD~\cite{gref2} split, which are denoted as RefCOCOg(g) and RefCOCOg(u), respectively.

To comprehensively evaluate the effectiveness and generalization capability of the proposed BRL method, we consider two experimental settings following recent works~\cite{barleria, detris}: The common setting evaluates dataset-specific adaptation capability, whereas the mixed setting further examines the robustness of multimodal models under diverse referring expressions and visual scenes.

\subsubsection{Metrics}

We employ two commonly used metrics, overall IoU (oIoU) and mean IoU (mIoU), in all experiments to quantify each method. The former computes the ratio between the aggregate intersection area and the aggregate union area across the entire dataset, which inherently favors larger objects. In contrast, the latter calculates the IoU for each sample individually and then averages across all samples, thereby treating objects of different sizes equally.

\subsection{Implementation Details}

We mainly use DINOv3~\cite{dinov3} pretrained ViT-B~\cite{vit} (DINOv3-B) with 12 layers and 768 hidden size, and ViT-L~\cite{vit} (DINOv3-L) with 24 layers and 1024 hidden size as the vision foundation model, combining the 12-layer language foundation model with hidden size 512 from CLIP-ViT-B~\cite{clip}, to evaluate the effectiveness of our method. For all experiments, the network is trained for 50 epochs using the Adam optimizer with a batch size of 32, a learning rate of 0.0001, and a step learning rate decay of 0.1 at epoch 35 on 8 RTX 4090 GPUs. Following previous works~\cite{cris, etris, detris}, the textual sentence length is set to 17 for RefCOCO, RefCOCO+, RefCOCOg(g), and the mixed dataset, and 22 for RefCOCOg(u). The task decoder has three layers with 8 heads of attention and a feed-forward hidden size of 512. Images are resized to 480$\times$480 without any data augmentations for training, and no additional post-processing operations are applied during inference. We keep the adapter dimension $d = 64$ as a default setting. Our BRL is applied at layers $ i\in[5, 8, 11]$ for both DINOv3-B and the text encoder, while at layers $ i\in[11, 17, 23]$ for DINOv3-L.

\subsection{Main Results}

We perform a comprehensive comparison with existing state-of-the-art (SOTA) methods under both the common and the mixed settings. The comparison results using the oIoU metric are reported in \cref{tab_ris_oiou}, while the mIoU comparisons are summarized in \cref{tab_ris_miou}. Existing methods are grouped into three categories, including traditional full fine-tuning~\cite{lavt, vlt, cgformer, rela, dmmi, refsegformer, magnet, remamber, hiea2g, cmanet, cmirnet, polyformer, vglaw, asda}, multimodal large language models (MLLMs)~\cite{lisa, M2SA, read, gsva, segllm, popen}, and parameter-efficient fine-tuning (PEFT)~\cite{etris, barleria, risclip, detris}.

\subsubsection{Comparison Under the Common Setting}

Under the single-dataset setting, our BRL consistently achieves superior performance across different benchmarks and backbone scales. Compared with existing PEFT methods, BRL demonstrates substantial gains on both oIoU and mIoU metrics. 

Specifically, using the DINOv2-B~\cite{dinov2} backbone, BRL-B surpasses DETRIS-B~\cite{detris} by +2.5\%/+1.5\%/+2.2\% oIoU on RefCOCO val/testA/testB and by +2.9\%/+2.6\%/+2.3\% mIoU, respectively. Similar improvements are consistently observed on RefCOCO+, where BRL-B achieves 73.7\% oIoU and 76.7\% mIoU on the testA split. On the more challenging RefCOCOg benchmark, BRL-B also achieves clear advantages over previous PEFT approaches, validating the effectiveness of reciprocal multimodal adaptation for long and complex referring expressions. Impressively, BRL-B outperforms DETRIS-L across all benchmarks except the RefCOCO+ testB split, highlighting the efficacy of our multimodal reciprocal learning. When equipped with the larger DINOv2-L backbone, BRL-L consistently improves all benchmarks across RefCOCO, RefCOCO+, and RefCOCOg datasets over DETRIS-L. In particular, BRL with DINOv3-L~\cite{dinov3} reaches 79.3\%, 81.3\%, and 75.4\% oIoU on the RefCOCO val, testA, and testB splits, respectively, while simultaneously obtaining 81.1\%, 82.4\%, and 77.7\% mIoU, setting new SOTA records. 

Moreover, BRL also achieves superior performance compared with recent fully fine-tuning RIS methods. Despite updating only lightweight reciprocal adapters, BRL with DINOv3-L achieves better results than fully fine-tuning methods such as CMANet~\cite{cmanet} and CMIRNet~\cite{cmirnet}. Notably, BRL-L under the common setting already surpasses the traditional full fine-tuning method CMIRNet-L trained under the mixed setting on RefCOCO. This observation demonstrates the remarkable efficiency of BRL, which achieves stronger performance using less training data and substantially lower optimization cost. Compared with large-scale MLLM-based models such as SegLLM~\cite{segllm} and POPEN~\cite{popen}, BRL achieves competitive performance while avoiding expensive multimodal instruction tuning and massive training data. These results demonstrate that effective bidirectional reciprocal adaptation can provide strong multimodal alignment without relying on heavy end-to-end optimization.

\begin{table*}[t]
\centering
\caption{Comparison with Existing SOTA Methods Using the \textbf{oIoU} Metric on the RefCOCO, RefCOCO+, and RefCOCOg Datasets. \\ $^\dagger$: Results are Produced Using the Mixed RefCOCO/+/g Datasets. $^\ddagger$: Results are Produced Using Large-scale Datasets. \\ The Best Results are in Bold.}
\label{tab_ris_oiou}
\resizebox{1.0\linewidth}{!}{
\begin{tabular}{l|c|cc|ccc|ccc|ccc}
\toprule
\multirow{2}{*}{Method} & \multirow{2}{*}{Venue} & \multicolumn{2}{c|}{Multimodal Backbone}           & \multicolumn{3}{c|}{RefCOCO~\cite{refcoco}} & \multicolumn{3}{c|}{RefCOCO+~\cite{refcoco}} & \multicolumn{3}{c}{RefCOCOg~\cite{gref1,gref2}}  \\ \cmidrule{3-13}
&                        & Vision            & Language           & val     & testA   & testB   & val     & testA    & testB   & val(u) & test(u) & val(g) \\ \midrule
\multicolumn{13}{c}{\textit{Traditional Full Fine-Tuning}}   \\ 
\midrule
LAVT~\cite{lavt}                    & CVPR'22                & Swin-B~\cite{swin}            & BERT-B~\cite{bert}             & 72.7    & 75.8    & 68.8    & 62.1    & 68.4     & 55.1    & 61.2    & 62.1     & 60.5  \\
VLT~\cite{vlt}                     & TPAMI'22               & Swin-B~\cite{swin}            & BERT-B~\cite{bert}             & 73.0    & 76.0    & 69.6    & 63.5    & 68.4     & 56.9    & 63.5    & 66.2     & 62.8   \\
CGFormer~\cite{cgformer}                & CVPR'23                & Swin-B~\cite{swin}            & BERT-B~\cite{bert}             & 74.8    & 77.3    & 70.6    & 64.5    & 71.0     & 57.1    & 64.7    & 65.1     & 62.5  \\
ReLA~\cite{rela}                    & CVPR'23                & Swin-B~\cite{swin}            & BERT-B~\cite{bert}             & 73.8    & 76.5    & 70.2    & 66.0    & 71.0     & 57.7    & 65.0    & 66.0     & 62.7  \\
DMMI~\cite{dmmi}                    & ICCV'23                & Swin-B~\cite{swin}            & BERT-B~\cite{bert}             & 74.1    & 77.1    & 70.2    & 64.0    & 69.7     & 57.0    & 63.5    & 64.2     & 62.0   \\
RefSegformer~\cite{refsegformer}            & TIP'24                 & Swin-B~\cite{swin}            & BERT-B~\cite{bert}             & 73.2    & 75.6    & 70.1    & 63.5    & 68.7     & 55.4    & 62.6    & 63.1     & 58.5   \\
MagNet~\cite{magnet}                  & CVPR'24                & Swin-B~\cite{swin}            & BERT-B~\cite{bert}             & 75.2    & 78.2    & 71.1    & 66.2    & 71.3     & 58.1    & 65.4    & 66.0     & 63.1  \\
ReMamber~\cite{remamber}                & ECCV'24                & VMamba-B~\cite{vmamba}          & BERT-B~\cite{bert}             & 74.5    & 76.7    & 70.9    & 65.0    & 70.8     & 57.5    & 63.9    & 64.0     & ---  \\
HieA2G~\cite{hiea2g}  & AAAI'25 & Swin-B~\cite{swin} & RoBERTa-B~\cite{roberta} & 75.1 & 77.6 & 71.1 & 66.5 & 71.4 & 58.9 & 65.3 & 66.6 & --- \\
CMANet~\cite{cmanet}  & TMM'25   & Swin-B~\cite{swin} & BERT-B~\cite{bert}    & \textbf{75.5} & \textbf{78.3} & \textbf{71.8} & \textbf{66.7} & \textbf{71.9} & \textbf{59.0} & \textbf{66.6} & \textbf{66.7} & \textbf{64.1} \\
CMIRNet~\cite{cmirnet} & TCSVT'25 & Swin-B~\cite{swin} & BERT-B~\cite{bert}    & 74.0 & 77.1 & 71.3 & 64.1 & 70.4 & 56.4 & 63.5 & 63.9 & ---   \\
\midrule

PolyFormer-B$^\dagger$~\cite{polyformer}            & CVPR'23                & Swin-B~\cite{swin}            & BERT-B~\cite{bert}             & 74.8    & 76.6    & 71.1    & 67.6    & 72.9     & 59.3    & 67.8    & 69.1     & ---      \\
MagNet$^\dagger$~\cite{magnet}                & CVPR'24                & Swin-B~\cite{swin}            & BERT-B~\cite{bert}             & 76.6    & 78.3    & 72.2    & 68.1    & 73.6     & 61.8    & 67.8    & 69.3     & ---    \\
CMANet$^\dagger$~\cite{cmanet}      & TMM'25   & Swin-B~\cite{swin}  & BERT-B~\cite{bert} & 77.3 & 80.2 & 73.1 & 68.2 & 73.8 & 60.4 & 67.4 & 70.1 & --- \\
CMIRNet-B$^\dagger$~\cite{cmirnet}    & TCSVT'25 & Swin-B~\cite{swin}  & BERT-B~\cite{bert} & 77.7 & 79.8 & 74.4 & 68.3 & 74.1 & 61.1 & 69.2 & 71.9 & --- \\
PolyFormer-L$^\dagger$~\cite{polyformer}            & CVPR'23                & Swin-L~\cite{swin}            & BERT-B~\cite{bert}             & 76.0    & 78.3    & 73.3    & 69.3    & \textbf{74.6}     & 61.9    & 69.2    & 70.2     & ---   \\
CMIRNet-L$^\dagger$~\cite{cmirnet}    & TCSVT'25 & Swin-L~\cite{swin}  & BERT-B~\cite{bert} & \textbf{78.2} & \textbf{80.4} & \textbf{75.2} & \textbf{69.8} & 73.5 & \textbf{62.1} & \textbf{70.4} & \textbf{72.1} & ---  \\
\midrule

\multicolumn{13}{c}{\textit{Multimodal Large Language Models}}  \\ 
\midrule
LISA-7B$^\ddagger$~\cite{lisa}               & CVPR'24                & \multicolumn{2}{c|}{SAM-H~\cite{sam} + LLaVA-7B~\cite{llava}} & 74.9    & 79.1    & 72.3    & 65.1    & 70.8     & 58.1    & 67.9    & 70.6     & ---  \\
M$^2$SA-7B$^\ddagger$~\cite{M2SA}               & ICLR'25                & \multicolumn{2}{c|}{SAM-H~\cite{sam} + LLaVA-7B~\cite{llava}} & 74.0    & 76.8    & 69.7    & 63.1    & 67.2     & 56.1    & 67.0    & 68.3     & ---   \\
READ-7B$^\ddagger$~\cite{read}               & CVPR'25                & \multicolumn{2}{c|}{SAM-H~\cite{sam} + LLaVA-7B~\cite{llava}} & 78.1    & 80.2    & 73.2    & 68.4    & 73.7     & 60.4    & 70.1    & 71.4     & ---  \\
GSVA-7B$^\ddagger$~\cite{gsva} & CVPR'24 & \multicolumn{2}{c|}{LLaVA-7B~\cite{llava}}  & 77.2    & 78.9    & 73.5    & 65.9    & 69.6     & 59.8    & 72.7    & 73.3     & ---  \\
SegLLM-7B$^\ddagger$~\cite{segllm}             & ICLR'25                & \multicolumn{2}{c|}{LLaVA-7B~\cite{llava}}          & \textbf{80.2}    & 81.5    & \textbf{75.4}    & 70.3    & 73.0     & 62.5    & 72.6    & 73.6     & ---  \\
POPEN-7B$^\ddagger$~\cite{popen} & CVPR'25  & \multicolumn{2}{c|}{LLaVA-7B~\cite{llava}}  & 79.3    & \textbf{82.0}    & 74.1    & \textbf{73.1}    & \textbf{77.0}     & \textbf{65.1}    & \textbf{75.4}    & \textbf{75.6}   & ---  \\
\midrule

\multicolumn{13}{c}{\textit{Parameter Efficient Fine-Tuning}}  \\ 
\midrule
RISCLIP~\cite{risclip}                 & NAACL'24               & \multicolumn{2}{c|}{CLIP-ViT-B~\cite{clip}}           & 73.6    & 76.5    & 69.8    & 65.5    & 70.6     & 55.5    & 64.1    & 65.1     & ---       \\
DETRIS-B~\cite{detris}                  & AAAI'25                & DINOv2-B~\cite{dinov2}          & CLIP-T-B~\cite{clip}           & 74.3    & 77.6    & 71.4    & 65.6    & 72.0     & 56.5    & 65.2    & 66.4     & 63.2   \\
BRL-B~(Ours)               & ---                & DINOv2-B~\cite{dinov2}          & CLIP-T-B~\cite{clip}           &  76.8   &  79.1   &  73.6  & 66.7  & 73.7  &  57.7  &  67.6   &  68.1    & 65.5  \\ 
\rowcolor[HTML]{E2F0D9}
BRL-B~(Ours)               & ---                & DINOv3-B~\cite{dinov3}          & CLIP-T-B~\cite{clip}           &  77.0   & 80.0    & 74.9   &  68.6   &  74.6   & 58.6  &  67.4   &  68.2    &  66.3   \\ 

DETRIS-L~\cite{detris}                  & AAAI'25                & DINOv2-L~\cite{dinov2}          & CLIP-T-B~\cite{clip}           & 76.1    & 78.3    & 73.2    & 67.9    & 73.3     & 60.2    & 66.8    & 68.0     & 65.0     \\
BRL-L~(Ours)               & ---                & DINOv2-L~\cite{dinov2}          & CLIP-T-B~\cite{clip}           &  78.0   &  80.3   &  74.7  & 69.6    & 74.7    & 61.5  &  69.2   &  \textbf{70.2}    &  67.3  \\ 
\rowcolor[HTML]{E2F0D9}
BRL-L~(Ours)                & ---                & DINOv3-L~\cite{dinov3}          & CLIP-T-B~\cite{clip}         &  \textbf{79.3}   & \textbf{81.3}    & \textbf{75.4}   &  \textbf{69.7}   &  \textbf{76.2}   & \textbf{62.3}   &  \textbf{69.5}   &  70.0    &  \textbf{68.0}  \\
\midrule

BRL-B$^\dagger$~(Ours)               & ---                & DINOv2-B~\cite{dinov2}          & CLIP-T-B~\cite{clip}           & 81.2    &  83.5   & 78.2   &  72.8   & 76.8    & 64.5  &  73.4   &  74.4    & ---    \\
\rowcolor[HTML]{E2F0D9}
BRL-B$^\dagger$~(Ours)               & ---                & DINOv3-B~\cite{dinov3}          & CLIP-T-B~\cite{clip}           &  81.7   &  83.2   &  78.3  &  73.1   &  78.1    &  65.5  &  73.6   &  75.8    &  ---  \\ 
DETRIS-L$^\dagger$~\cite{detris}                & AAAI'25                & DINOv2-L~\cite{dinov2}          & CLIP-T-B~\cite{clip}           & 80.7    & 82.2    & 77.9    & 73.1    & 77.5     & 66.0    & 73.2    & 74.7     & ---    \\
BRL-L$^\dagger$~(Ours)               & ---                & DINOv2-L~\cite{dinov2}          & CLIP-T-B~\cite{clip}           & 82.2    &  84.2   & 79.4   & \textbf{74.7}    &  78.8   & 66.7  & 74.8    & 76.1     & ---   \\
\rowcolor[HTML]{E2F0D9}
BRL-L$^\dagger$~(Ours)               & ---                & DINOv3-L~\cite{dinov3}          & CLIP-T-B~\cite{clip}           &  \textbf{82.4}   &  \textbf{84.6}    & \textbf{79.8}   &  74.6   &  \textbf{79.6}   & \textbf{67.3}  & \textbf{75.5}    &  \textbf{76.7}    &  ---  \\
\bottomrule           
\end{tabular}}
\end{table*}

\subsubsection{Comparison Under the Mixed Setting}

Under the mixed-dataset setting, our BRL exhibits stronger generalization capability. Benefiting from reciprocal multimodal learning across diverse referring distributions, BRL consistently outperforms existing methods on all benchmarks.

Specifically, BRL with DINOv3-B~\cite{dinov3} achieves 81.7\%, 78.1\%, and 75.8\% oIoU on RefCOCO val, RefCOCO+ testA, and RefCOCOg val(u), respectively, outperforming previous PEFT methods by clear margins. In terms of mIoU, BRL-B further achieves 82.4\%, 80.4\%, and 77.3\% on the corresponding benchmarks, validating the robustness of the proposed reciprocal adaptation mechanism. More importantly, BRL with DINOv2-B already surpasses DETRIS~\cite{detris} equipped with the stronger DINOv2-L backbone on the RefCOCO benchmark under both oIoU and mIoU metrics. For example, BRL-B achieves 81.2\%/83.5\%/78.2\% oIoU while DETRIS-L gets 80.7\%/82.2\%/77.9\% oIoU on the RefCOCO val/testA/testB splits, clearly exceeding the latter despite using a smaller backbone. This further verifies the effectiveness and scalability of the proposed bidirectional reciprocal learning scheme.

In addition, BRL with DINOv3-L establishes new state-of-the-art PEFT performance across almost all datasets and metrics. Specifically, it achieves 82.4\%, 84.6\%, and 79.8\% oIoU on RefCOCO val/testA/testB, together with 83.5\%, 85.0\%, and 81.0\% mIoU. Compared with DETRIS-L under the same setting, BRL-L consistently improves both localization accuracy and segmentation quality across all RIS benchmarks. Moreover, BRL also surpasses all fully fine-tuning methods, despite requiring significantly fewer trainable parameters and computational overhead.

Overall, the experimental results demonstrate that BRL achieves highly effective multimodal adaptation under frozen foundation backbones. By jointly modeling token-level reciprocal interaction and channel-level reciprocal modulation, the proposed framework establishes stronger bidirectional vision-language alignment, leading to consistent improvements across different datasets, metrics, and backbone scales.

\begin{table*}[t]
\centering
\caption{Comparison with Existing SOTA Methods Using the \textbf{mIoU} Metric on the RefCOCO, RefCOCO+, and RefCOCOg Datasets. \\ $^\dagger$: Results are Produced Using the Mixed RefCOCO/+/g Datasets.}
\label{tab_ris_miou}
\resizebox{1.0\linewidth}{!}{
\begin{tabular}{l|c|cc|ccc|ccc|ccc}
\toprule
\multirow{2}{*}{Method} & \multirow{2}{*}{Venue} & \multicolumn{2}{c|}{Multimodal Backbone} & \multicolumn{3}{c|}{RefCOCO~\cite{refcoco}} & \multicolumn{3}{c|}{RefCOCO+~\cite{refcoco}} & \multicolumn{3}{c}{RefCOCOg~\cite{gref1,gref2}}  \\ \cmidrule{3-13}
&                        & Vision         & Language    & val     & testA   & testB   & val     & testA    & testB   & val(u) & test(u) & val(g)  \\ \midrule
\multicolumn{13}{c}{\textit{Traditional Full Fine-Tuning}} \\ 
\midrule
LAVT~\cite{lavt}                    & CVPR'22                & Swin-B~\cite{swin}         & BERT-B~\cite{bert}      & 74.5    & 76.9    & 70.9    & 65.8    & 71.0     & 59.2    & 63.3    & 63.6     & 63.7    \\
CGFormer~\cite{cgformer}                & CVPR'23                & Swin-B~\cite{swin}         & BERT-B~\cite{bert}      & \textbf{76.9}    & \textbf{78.7}    & \textbf{73.3}    & 68.6    & 73.8     & 61.7    & 67.6    & 67.8     & 65.8   \\
ReLA~\cite{rela}                    & CVPR'23                & Swin-B~\cite{swin}         & BERT-B~\cite{bert}      & 75.6    & 77.8    & 72.8    & \textbf{70.4}    & \textbf{74.8}     & \textbf{63.9}    & \textbf{68.7}    & \textbf{69.6}     & \textbf{66.9}  \\
VG-LAW~\cite{vglaw}  & CVPR'23  & ViTDet-B~\cite{vitdet}  & BERT-B~\cite{bert} & 75.6 & 77.5 & 72.9 & 66.6 & 70.4 & 58.9 & 65.6 & 66.1 & ---  \\
ASDA~\cite{asda}                    & ACMMM'24               & \multicolumn{2}{c|}{CLIP-ViT-B~\cite{clip}}    & 75.1    & 77.1    & 71.4    & 66.8    & 71.1     & 57.8    & 65.7    & 66.5     & 63.6     \\
CMIRNet~\cite{cmirnet}     & TCSVT'25 & Swin-B~\cite{swin} & BERT-B~\cite{bert} & 76.0 & 78.1 & 73.1 & 68.2 & 73.0 & 61.3 & 66.7 & 66.5 & --- \\
\midrule

PolyFormer-B$^\dagger$~\cite{polyformer}  & CVPR'23  & Swin-B~\cite{swin}  & BERT-B~\cite{bert} & 76.0 & 77.1 & 73.2 & 70.7 & 74.5 & 64.6 & 69.4 & 69.9 & ---  \\
CMIRNet-B$^\dagger$~\cite{cmirnet}    & TCSVT'25 & Swin-B~\cite{swin} & BERT-B~\cite{bert} & 78.5 & 80.2 & 76.0 & 71.2 & 75.3 & 65.0 & 71.5 & 72.8 & ---  \\
PolyFormer-L$^\dagger$~\cite{polyformer}  & CVPR'23  & Swin-L~\cite{swin}  & BERT-B~\cite{bert} & 76.9 & 78.5 & 74.8 & 72.2 & 75.7 & 66.7 & 71.2 & 71.2 & ---  \\
CMIRNet-L$^\dagger$~\cite{cmirnet}    & TCSVT'25 & Swin-L~\cite{swin} & BERT-B~\cite{bert} & \textbf{79.0} & \textbf{80.7} & \textbf{76.6} & \textbf{72.5} & \textbf{77.0} & \textbf{66.9} & \textbf{72.6} & \textbf{73.2} & ---  \\
\midrule

\multicolumn{13}{c}{\textit{Parameter Efficient Fine-Tuning}} \\ 
\midrule
ETRIS~\cite{etris}                   & ICCV'23                & \multicolumn{2}{c|}{CLIP-ViT-B~\cite{clip}}    & 70.5    & 73.5    & 66.6    & 60.1    & 66.9     & 50.2    & 59.8    & 59.9     & 57.9        \\
BarLeRIa-B~\cite{barleria}                & ICLR'24                & \multicolumn{2}{c|}{CLIP-ViT-B~\cite{clip}}    & 72.4    & 75.9    & 68.3    & 65.0    & 70.8     & 56.9    & 63.4    & 63.8     & 61.6        \\
RISCLIP~\cite{risclip}                 & NAACL'24               & \multicolumn{2}{c|}{CLIP-ViT-B~\cite{clip}}    & 75.7    & 78.0    & 72.5    & 69.2    & 73.5     & 60.7    & 67.6    & 68.0     & ---     \\
DETRIS-B~\cite{detris}                  & AAAI'25                & DINOv2-B~\cite{dinov2}       & CLIP-T-B~\cite{clip}    & 76.0    & 78.2    & 73.5    & 68.9    & 74.0     & 61.5    & 67.9    & 68.1     & 65.9           \\
BRL-B~(Ours)               & ---                & DINOv2-B~\cite{dinov2}          & CLIP-T-B~\cite{clip}           &  78.9   &  80.8   &  75.8  &  71.1   & 76.7    & 63.8  &  70.7   &  71.2    &  68.5  \\
\rowcolor[HTML]{E2F0D9}
BRL-B~(Ours)               & ---                & DINOv3-B~\cite{dinov3}          & CLIP-T-B~\cite{clip}           &  79.3   &  81.4   & 77.1   &  72.1   &  77.2   &  63.7  &  71.2   &  71.5    &  70.2   \\

BarLeRIa-L~\cite{barleria}  & ICLR'24  & \multicolumn{2}{c|}{EVA-CLIP-ViT-L~\cite{eva_clip}} & 76.8 & 79.0 & 74.0 & 71.5 & 76.2 & 65.4 & 68.7 & 69.7 & ---  \\
DETRIS-L~\cite{detris}                  & AAAI'25                & DINOv2-L~\cite{dinov2}       & CLIP-T-B~\cite{clip}    & 77.3    & 79.0    & 75.2    & 70.8    & 75.3     & 64.7    & 69.3    & 70.2     & 67.9           \\
BRL-L~(Ours)               & ---                & DINOv2-L~\cite{dinov2}          & CLIP-T-B~\cite{clip}           &  80.2   &  82.1   &  77.6  &  73.5   & 77.6    &  66.8 & 72.8    &  \textbf{73.3}    &  71.0 \\
\rowcolor[HTML]{E2F0D9}
BRL-L~(Ours)                & ---                & DINOv3-L~\cite{dinov3}          & CLIP-T-B~\cite{clip}         &  \textbf{81.1}   & \textbf{82.4}    & \textbf{77.7}   &  \textbf{73.9}   &  \textbf{78.6}   &  \textbf{67.3}  &  \textbf{73.4}   &   73.2   &  \textbf{71.9}   \\ 
\midrule

BarLeRIa-B$^\dagger$~\cite{barleria} & ICLR'24  & \multicolumn{2}{c|}{EVA-CLIP-ViT-B~\cite{eva_clip}}     & 77.6 & 79.4 & 75.3 & 71.7 & 75.7 & 66.0 & 70.9 & 71.4 & ---    \\
RISCLIP$^\dagger$~\cite{risclip}                & NAACL'24               & \multicolumn{2}{c|}{CLIP-ViT-B~\cite{clip}}    & 76.0    & 78.6    & 71.9    & 69.7    & 74.3     & 61.4    & 69.6    & 69.6     & ---         \\
BRL-B$^\dagger$~(Ours)               & ---                & DINOv2-B~\cite{dinov2}          & CLIP-T-B~\cite{clip}           &  82.2   &  83.9   & 79.7   &  75.9   & 79.4    & 69.7  &  75.5   &  76.3    & ---     \\
\rowcolor[HTML]{E2F0D9}
BRL-B$^\dagger$~(Ours)               & ---                & DINOv3-B~\cite{dinov3}          & CLIP-T-B~\cite{clip}           &  82.4   & 83.8    &  79.4  &  76.3   &  80.4   &  70.2  &  76.0  &  77.3  & ---    \\ 

BarLeRIa-L$^\dagger$~\cite{barleria} & ICLR'24  & \multicolumn{2}{c|}{EVA-CLIP-ViT-L~\cite{eva_clip}} & 79.0 & 80.8 & 77.0 & 74.2 & 77.8 & 68.3 & 72.7 & 73.3 & ---    \\
DETRIS-L$^\dagger$~\cite{detris}                 & AAAI'25                & DINOv2-L~\cite{dinov2}       & CLIP-T-B~\cite{clip}    & 81.0    & 81.9    & 79.0    & 75.2    & 78.6     & 70.2    & 74.6    & 75.3     & ---     \\
BRL-L$^\dagger$~(Ours)               & ---                & DINOv2-L~\cite{dinov2}          & CLIP-T-B~\cite{clip}           &  83.2   & 84.6    & 80.8   & \textbf{77.8}    &  81.2   & 71.2  &  76.6   &  77.9    & ---    \\
\rowcolor[HTML]{E2F0D9}
BRL-L$^\dagger$~(Ours)               & ---                & DINOv3-L~\cite{dinov3}          & CLIP-T-B~\cite{clip}           &  \textbf{83.5}   & \textbf{85.0}    & \textbf{81.0}   & 77.7    & \textbf{81.4}    & \textbf{71.8}  & \textbf{77.8}    &  \textbf{78.3}    &  ---   \\
\bottomrule
\end{tabular}}
\end{table*}

\subsection{Ablation Study}

We conduct extensive ablation studies on RefCOCO to validate the effectiveness of our BRL. Unless otherwise specified, all experiments are performed using DINOv3~\cite{dinov3} with CLIP~\cite{clip} text encoder under the common setting. In addition, we further investigate the compatibility and robustness of BRL across different vision foundation models (VFMs)~\cite{dino, mae, eva2, dinov2} with varying architectures and model scales.

\subsubsection{Comparison with Other PEFT Methods}

\begin{table}[t]
\centering
\setlength{\tabcolsep}{3.5pt}
\caption{Comparison with Existing PEFT Methods Using DINOv3-B/L~\cite{dinov3} Visual Encoder with CLIP-ViT-B~\cite{clip} Linguistic Encoder as the Vision-Language Model on RefCOCO.}
\label{tab_abla_sota}
\resizebox{\linewidth}{!}{
\begin{tabular}{l|ccc|ccc|rr}
\toprule
\multirow{2}{*}{PEFT Method} & \multicolumn{3}{c|}{RefCOCO (oIoU)}            & \multicolumn{3}{c|}{RefCOCO (mIoU)}  & \multicolumn{2}{c}{Tunable Cost} \\ \cmidrule{2-9}
& val           & testA         & testB         & val           & testA         & testB             & Params.         & FLOPs          \\
\midrule

FFT       & 65.7          & 68.7          & 61.8          & 69.9          & 71.9          & 66.7        & 149.1M          & 78.1G         \\
Fix Encoder             & 74.0          & 76.5          & 70.2          & 75.8          & 77.8          & 72.6                   & 0.0M            & 0.0G           \\
Adapter~\cite{adapter}                & 74.7          & 76.8          & 71.7          & 76.4          & 78.1          & 73.9          & 1.3M            & 0.5G           \\
AdaptFormer~\cite{adaptformer}                          & 74.6          & 77.1          & 71.2          & 76.6          & 78.3          & 73.9           & 1.3M            & 0.5G           \\
ETRIS~\cite{etris}                     & 74.9          & 77.4          & 71.0          & 76.5          & 78.3          & 73.4              & 1.4M            & 0.7G           \\
DETRIS~\cite{detris}               & 75.3          & 78.0          & 72.1          & 77.0          & 79.0          & 74.6               & 2.7M            & 1.9G           \\
\rowcolor[HTML]{E2F0D9}
BRL-B (Ours)                         & \textbf{77.0} & \textbf{80.0} & \textbf{74.9} & \textbf{79.3} & \textbf{81.4} & \textbf{77.1}     & 1.5M            & 1.4G           \\
BRL-B w/FFT                 & 68.6          & 70.7          & 64.9          & 72.7          & 74.1          & 69.6        & 150.6M          & 79.5G         \\
\midrule

FFT            & 66.8          & 69.2          & 63.0          & 70.3          & 72.2          & 68.1                 & 366.6M          & 274.9G         \\
Fix Encoder                  & 75.3          & 78.3          & 71.9          & 77.0          & 79.3          & 74.2                    & 0.0M            & 0.0G           \\
\rowcolor[HTML]{E2F0D9}
BRL-L (Ours)     & \textbf{79.3}          & \textbf{81.3}          & \textbf{75.4}          & \textbf{81.1}          & \textbf{82.4}          & \textbf{77.7}                   & 1.8M            & 1.7G           \\
\bottomrule
\end{tabular}}
\end{table}

To comprehensively evaluate the effectiveness of BRL, we first compare it with representative PEFT methods~\cite{adapter, adaptformer, etris,detris}. As reported in \cref{tab_abla_sota}, BRL consistently achieves the best performance while maintaining extremely low tunable costs.

Under the setting of using DINOv3-B, BRL achieves 77.0\%/80.0\%/74.9\% oIoU and 79.3\%/81.4\%/77.1\% mIoU on the RefCOCO val/testA/testB splits, surpassing previous PEFT methods by large margins across all benchmarks. In particular, BRL improves DETRIS~\cite{detris} by +1.7\%, +2.0\%, and +2.8\% oIoU on the three splits, respectively, while requiring only 1.5M trainable parameters and 1.4G tunable FLOPs. The impressive results indicate that reciprocal bidirectional interaction is substantially more effective than unidirectional adaptation. It is also worth noting that directly applying full fine-tuning (FFT) to VFMs results in severe performance degradation. Specifically, FFT only achieves 65.7 oIoU and 69.9 mIoU on RefCOCO val, which is dramatically lower than both the fix encoder and PEFT-based adaptation. Similar trends can also be observed under the DINOv3-L backbone. This suggests that large-scale foundation models with strong pre-trained knowledge are highly sensitive to full-parameter optimization, making them prone to catastrophic forgetting. In contrast, PEFT methods preserve the majority of parameters and only optimize a small number of lightweight modules, thereby achieving efficiency. Among them, BRL further achieves superior performance by introducing reciprocal bidirectional adaptation, enabling efficient task-specific learning while maximally preserving the strong prior knowledge of frozen vision and language foundation models.

\begin{figure*}[t]
    \centering
    \includegraphics[width=1\linewidth]{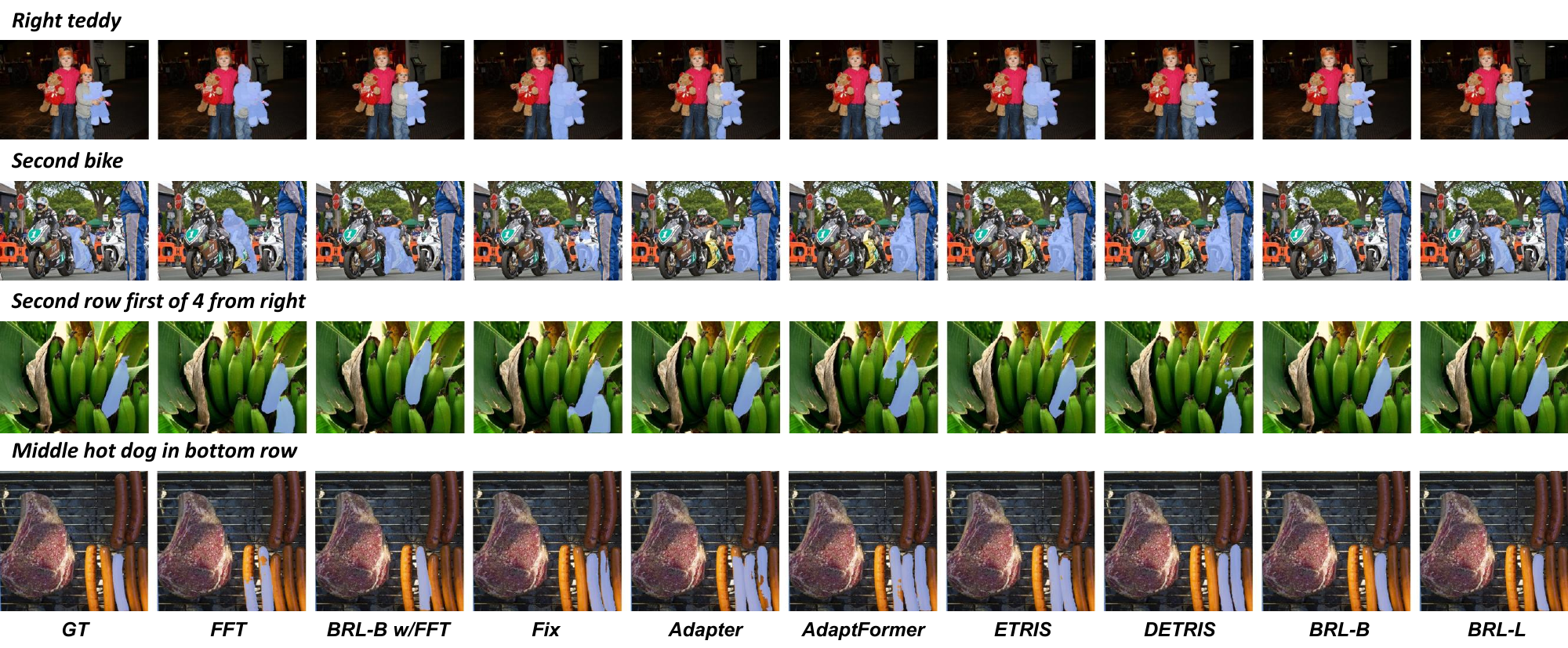}
    \caption{Qualitative comparisons of BRL with other PEFT methods using DINOv3~\cite{dinov3}.}
    \label{fig_peft_comp}
\end{figure*}

As illustrated in \cref{fig_peft_comp}, existing PEFT methods often suffer from incomplete object localization or distracted attention regions due to insufficient cross-modal interaction. By contrast, BRL produces substantially more accurate and semantically consistent segmentation results, especially for challenging scenarios involving ambiguous expressions, small objects, and complex backgrounds. This improvement mainly benefits from the cooperative design of reciprocal token interaction and reciprocal channel modulation, which jointly enhance fine-grained spatial grounding and global semantic consistency. 

\subsubsection{Effect of BRL's Core Components}

\begin{table}[t]
\centering
\setlength{\tabcolsep}{4pt}
\caption{Ablation Study on the Core Components of Our BRL. RAA: Reciprocal Attention Adapter. RGA: Reciprocal Gate Adapter.}
\label{tab_abla_comp}
\resizebox{\linewidth}{!}{
\begin{tabular}{cc|ccc|ccc|rr}
\toprule
\multirow{2}{*}{RAA} & \multirow{2}{*}{RGA} & \multicolumn{3}{c|}{RefCOCO (oIoU)}            & \multicolumn{3}{c|}{RefCOCO (mIoU)}            & \multicolumn{2}{c}{Tunable Cost} \\ \cmidrule{3-10}
                     &                      & val           & testA         & testB         & val           & testA         & testB         & Params.          & FLOPs         \\
\midrule
\ding{55}                   & \ding{55}                   & 74.0          & 76.5          & 70.2          & 75.8          & 77.8          & 72.6          & 0.0M             & 0.0G          \\
\ding{51}                   & \ding{55}                   & 75.9          & 78.6          & 73.2          & 78.2          & 80.3          & 75.5          & 0.8M             & 1.0G          \\
\ding{55}                   & \ding{51}                   & 75.3          & 78.8          & 72.9          & 78.0          & 80.1          & 75.4          & 0.7M             & 0.4G          \\
\rowcolor[HTML]{E2F0D9}
\ding{51}                    & \ding{51}                    & \textbf{77.0} & \textbf{80.0} & \textbf{74.9} & \textbf{79.3} & \textbf{81.4} & \textbf{77.1} & 1.5M             & 1.4G           \\ 
\bottomrule                 
\end{tabular}}
\end{table}

We further investigate the effectiveness of the two core components in BRL, namely the Reciprocal Attention Adapter (RAA) and the Reciprocal Gate Adapter (RGA), as reported in \cref{tab_abla_comp}.

\begin{figure}[t]
    \centering
    \includegraphics[width=1\linewidth]{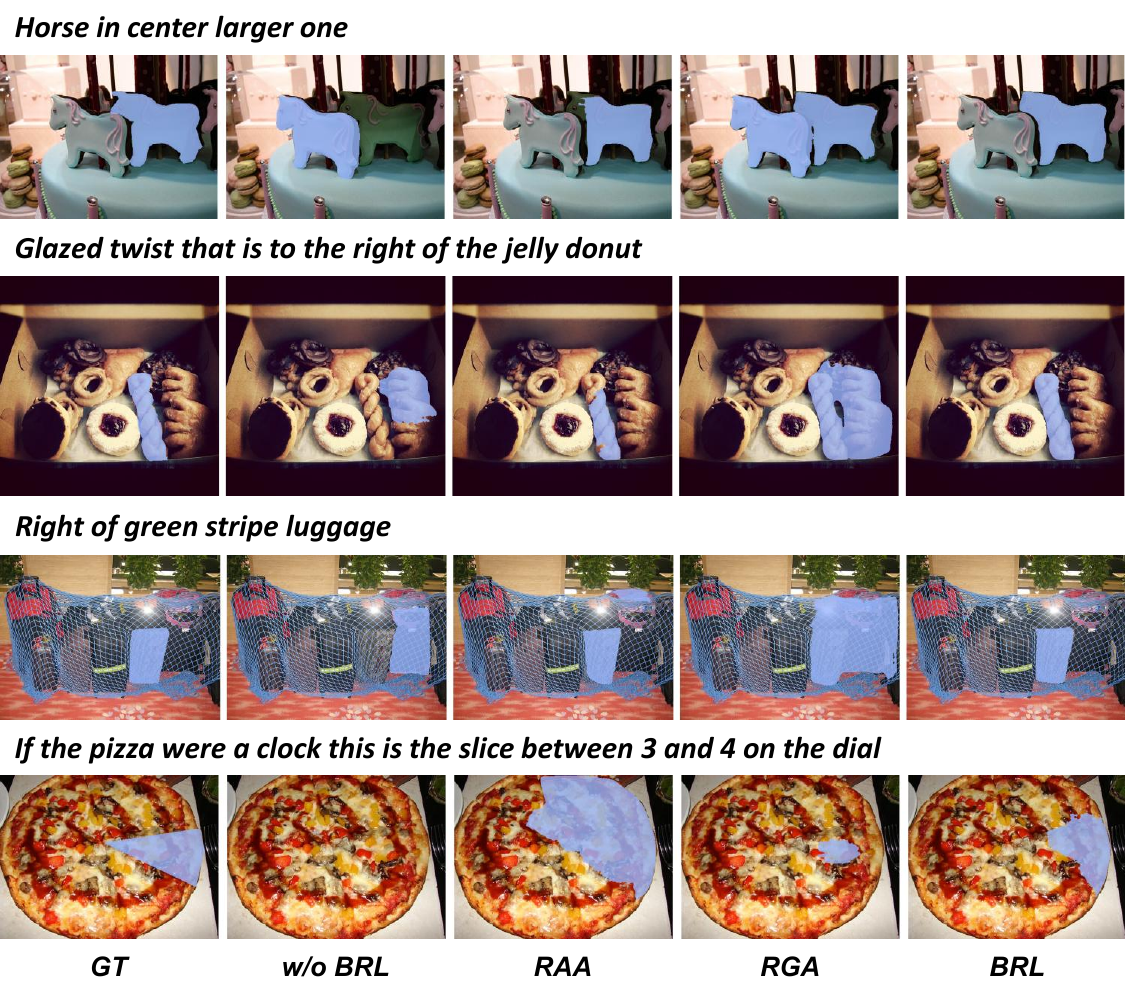}
    \caption{Qualitative results on the core components of BRL.}
    \label{fig_comp}
\end{figure}

Starting from the frozen baseline without any reciprocal adaptation modules, the model achieves 74.0\%/76.5\%/70.2\% oIoU and 75.8\%/77.8\%/72.6\% mIoU on the RefCOCO val/testA/testB splits, respectively. After introducing the RAA module, the performance is consistently improved by +1.9\%/+2.1\%/+3.0\% oIoU and +2.4\%/+2.5\%/+2.9\% mIoU across the three splits. These gains demonstrate that reciprocal token-level interaction effectively enhances fine-grained cross-modal correspondence learning by establishing bidirectional spatial alignment between visual and linguistic tokens. 

Similarly, equipping the model solely with the RGA module also yields notable improvements over the baseline, bringing gains of +1.3\%/+2.3\%/+2.7\% oIoU and +2.2\%/+2.3\%/+2.8\% mIoU, respectively. Compared with RAA, RGA introduces slightly fewer parameters and FLOPs while still producing strong improvements, indicating that channel-level reciprocal modulation plays an important role in enhancing global semantic consistency and adaptive feature selection.

When RAA and RGA are jointly employed, BRL achieves the best overall performance, improving the frozen model by +3.0\%/+3.5\%/+4.7\% oIoU and +3.5\%/+3.6\%/+4.5\% mIoU, respectively. The greater gains demonstrate strong complementarity between reciprocal token interaction and reciprocal channel regulation. Moreover, despite only introducing 1.5M trainable parameters and 1.4G tunable FLOPs, our BRL is capable of making remarkable performance, further demonstrating the efficiency of the proposed reciprocal adaptation.

\begin{figure*}[t]
    \centering
    \includegraphics[width=1\linewidth]{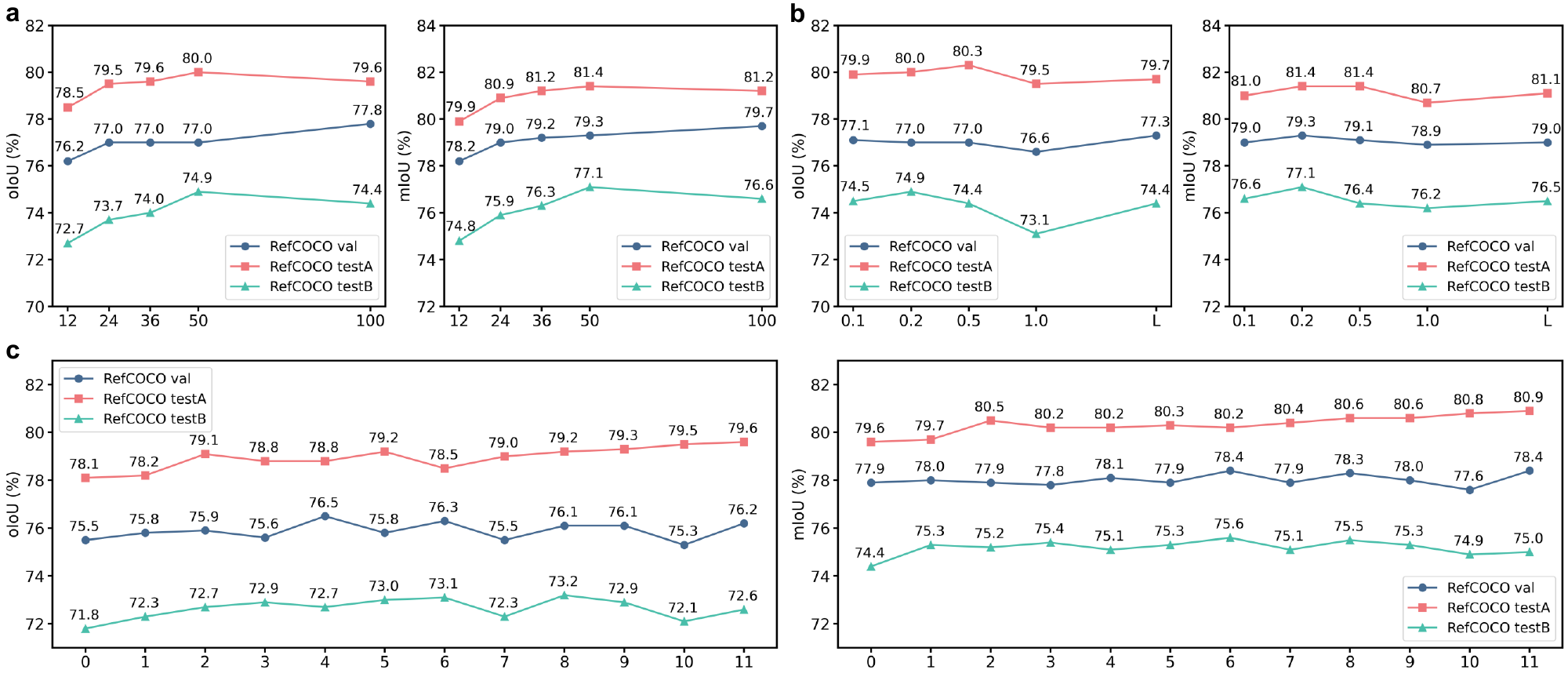}
    \caption{Ablation results on hyperparameter settings of our BRL on RefCOCO. \arial{\textbf{a}}: Performance comparison across different training epochs, where we adopt 50 epochs for optimal efficiency-accuracy trade-off. \arial{\textbf{b}}: Sensitivity analysis of the residual scaling factor $\alpha$ for controlling the reciprocal adaptation strength, with \arial{L} denoting learnable parameters. We select $\alpha=0.2$ as it achieves the best balance. \arial{\textbf{c}}: Layer-wise ablation study of inserting RAA and RGA into individual Transformer layers, demonstrating the impact of adapter placement on model performance.}
    \label{fig_hyper}
\end{figure*}

\begin{table}[t]
\centering
\setlength{\tabcolsep}{4.5pt}
\caption{Ablation Study on the Adapter Dimension $d$ of Our BRL.}
\label{tab_abla_dim}
\resizebox{\linewidth}{!}{
\begin{tabular}{c|ccc|ccc|rr}
\toprule
\multirow{2}{*}{$d$} & \multicolumn{3}{c|}{RefCOCO (oIoU)}            & \multicolumn{3}{c|}{RefCOCO (mIoU)}            & \multicolumn{2}{c}{Tunable Cost} \\ \cmidrule{2-9}
                     & val           & testA         & testB         & val           & testA         & testB         & Params.         & FLOPs          \\ \midrule
16   &  76.5  &  79.3          & 74.2          & 78.9          & 80.9          & 76.4          & 0.3M   & 0.7G  \\
32   & 77.1          & 80.0          & 74.6          & 79.0          & 81.1          & 76.9          & 0.7M            & 0.9G           \\
\rowcolor[HTML]{E2F0D9}
64                   & 77.0          & 80.0          & \textbf{74.9} & 79.3          & 81.4          & 77.1 & 1.5M            & 1.4G           \\
128                  & \textbf{77.5}          & \textbf{80.2} & 74.7          & \textbf{79.6} & \textbf{81.6} & \textbf{77.4}          & 3.3M            & 2.5G           \\
256                  & 77.4          & 79.5          & 74.5          & 79.6          & 81.1          & 77.0          & 7.9M            & 5.1G     \\ \bottomrule           
\end{tabular}}
\end{table}

As illustrated in \cref{fig_comp}, removing either RAA or RGA leads to noticeable degradation in segmentation quality. In particular, models without RAA tend to produce inaccurate object localization due to insufficient fine-grained cross-modal interaction, while removing RGA often causes semantic confusion and incomplete object regions. By jointly modeling reciprocal token interaction and reciprocal channel modulation, BRL generates substantially more accurate and semantically consistent segmentation masks under challenging referring scenarios.

\begin{table}[t]
\centering
\setlength{\tabcolsep}{3pt}
\caption{Ablation Study of Different VFMs with BRL on RefCOCO.}
\label{tab_abla_vfms}
\resizebox{\linewidth}{!}{
\begin{tabular}{l|ccc|ccc|rr}
\toprule
\multirow{2}{*}{VFMs} & \multicolumn{3}{c|}{RefCOCO (oIoU)} & \multicolumn{3}{c|}{RefCOCO (mIoU)} & \multicolumn{2}{c}{Fix Cost} \\ \cmidrule{2-9}
                      & val       & testA      & testB     & val       & testA      & testB     & Params.            & FLOPs             \\
\midrule

\multicolumn{9}{l}{\textit{Tunable Cost (Params. / FLOPs): 1.1M / 1.1G}}  \\ 
\midrule
DINOv2-S~\cite{dinov2}              & 74.5      & 76.8       & 70.7      & 76.5      & 78.3       & 72.7      & 85.5M         & 32.6G        \\
\rowcolor[HTML]{E2F0D9}
DINOv3-S+~\cite{dinov3}             & \textbf{75.4}      & \textbf{79.5}       & \textbf{72.5}      & \textbf{77.8}      & \textbf{80.3}       & \textbf{75.1}      & 92.1M         & 26.6G        \\
\midrule

\multicolumn{9}{l}{\textit{Tunable Cost (Params. / FLOPs): 1.5M / 1.4G}}  \\ 
\midrule
DINO-B~\cite{dino}                & 69.1      & 71.1       & 64.7      & 71.4      & 73.2       & 67.4      & 149.2M        & 92.8G        \\
MAE-B~\cite{mae}                 & 73.0      & 75.3       & 69.4      & 75.9      & 77.6       & 72.4       & 149.2M        & 92.8G        \\
EVA2-B~\cite{eva2}                & 74.8      & 78.1       & 71.6      & 76.3      & 78.8       & 73.5      & 149.8M        & 92.9G        \\
DINOv2-B~\cite{dinov2}              & 76.8      & 79.1       & 73.6      & 78.9      & 80.8       & 75.8      & 150.0M        & 108.2G       \\
\rowcolor[HTML]{E2F0D9}
DINOv3-B~\cite{dinov3}              & \textbf{77.0}      & \textbf{80.0}       & \textbf{74.9}      & \textbf{79.3}      & \textbf{81.4}       & \textbf{77.1}      & 149.1M        & 78.1G        \\
\midrule

\multicolumn{9}{l}{\textit{Tunable Cost (Params. / FLOPs): 1.8M / 1.7G}}  \\ 
\midrule
MAE-L~\cite{mae}                 & 76.5      & 78.5       & 72.8      & 79.1      & 80.6       & 75.9      & 366.7M        & 313.6G       \\
EVA2-L~\cite{eva2}                & 79.3      & \textbf{82.1}       & \textbf{75.6}      & 80.6      & \textbf{82.8}       & 77.4      & 367.5M        & 313.9G       \\
DINOv2-L~\cite{dinov2}              & 78.0      & 80.3       & 74.7      & 80.2      & 82.1       & 77.6      & 367.8M        & 364.4G       \\
\rowcolor[HTML]{E2F0D9}
DINOv3-L~\cite{dinov3}              & \textbf{79.3}      & 81.3       & 75.4      & \textbf{81.1}      & 82.4       & \textbf{77.7}      & 366.6M        & 274.9G       \\
\midrule
\multicolumn{9}{l}{\textit{Tunable Cost (Params. / FLOPs): 2.0M / 1.9G (DINOv2-G: 2.3M / 2.1G)}}  \\ 
\midrule
MAE-H~\cite{mae}                 & 77.2      & 79.0       & 71.8      & 79.8      & 81.4       & 75.5      & 694.2M        & 839.7G       \\
DINOv2-G~\cite{dinov2}              & 78.9      & 81.0       & 75.3      & 80.9      & 82.5       & 77.9      & 1199.9M       & 1297.7G      \\
\rowcolor[HTML]{E2F0D9}
DINOv3-H+~\cite{dinov3}             & \textbf{80.0}      & \textbf{82.2}       & \textbf{76.3}      & \textbf{82.0}      & \textbf{83.4}       & \textbf{78.5}      & 904.1M        & 761.1G        \\ 
\bottomrule    
\end{tabular}}
\end{table}

\subsubsection{Effect of Hyperparameter Settings}

We further conduct detailed studies on several important hyperparameters of BRL, including the training epochs, residual scaling factor $\alpha$, adapter insertion layers, and adapter bottleneck dimension $d$.

\paragraph{Training Epochs}

We first investigate the influence of training epochs. As illustrated in \cref{fig_hyper}.\arial{\textbf{a}}, the performance improves steadily with longer training and gradually saturates around 50 epochs. Notably, BRL already achieves 76.2\%/78.5\%/72.7\% oIoU and 78.2\%/79.9\%/74.8\% mIoU after only 12 epochs, which already surpasses most existing PEFT methods trained with substantially longer schedules, indicating the fast convergence and high optimization efficiency of our BRL. Increasing the training epochs to 50 further improves the performance to 77.0\%/80.0\%/74.9\% oIoU and 79.3\%/81.4\%/77.1\% mIoU. Although training for 100 epochs slightly improves the validation results, the test performance begins to fluctuate, indicating potential overfitting. Therefore, we adopt 50 epochs as the default setting to achieve a favorable balance between performance and computational cost.

\begin{figure*}[t]
    \centering
    \includegraphics[width=1\linewidth]{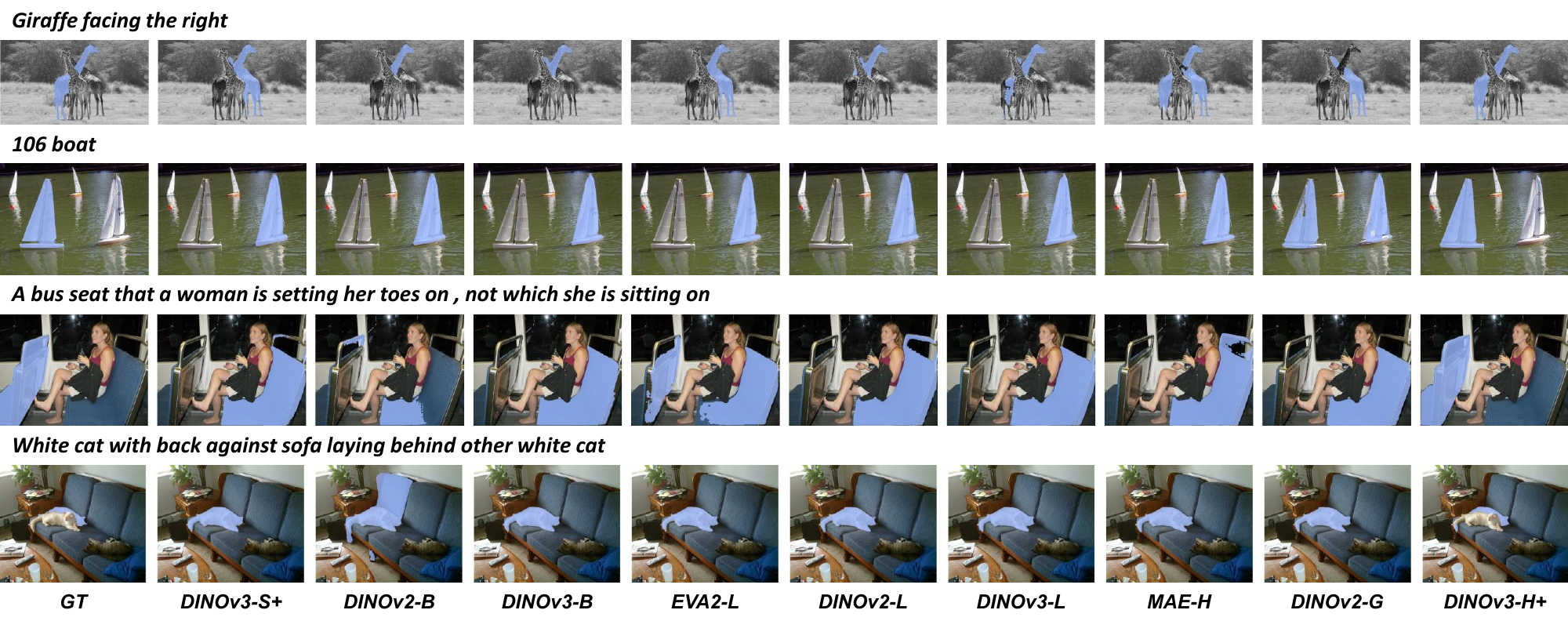}
    \caption{Qualitative comparison of BRL with different vision foundation models~\cite{mae, eva2, dinov2, dinov3} under the RIS mixed benchmark setting.}
    \label{fig_mix}
\end{figure*}

\paragraph{Residual Scaling Factor $\alpha$}

We further analyze the influence of $\alpha$, which controls the strength of reciprocal adaptation. As shown in \cref{fig_hyper}.\arial{\textbf{b}}, BRL remains relatively stable across different settings, demonstrating good robustness. Moderate scaling factors generally perform better, while overly large values may disturb the frozen representations. Specifically, $\alpha=0.2$ achieves the best overall performance with 77.0\%/80.0\%/74.9\% oIoU and 79.3\%/81.4\%/77.1\% mIoU. In contrast, $\alpha=1.0$ causes noticeable performance degradation, especially on the testB split. We also evaluate a learnable scaling strategy, which achieves competitive but not superior results compared with fixed moderate values.

\paragraph{Adapter Insertion Layers}

To explore where reciprocal adaptation is most effective, we insert the RAA and RGA modules individually into different transformer layers. As illustrated in \cref{fig_hyper}.\arial{\textbf{c}}, BRL achieves stable improvements across nearly all insertion positions, demonstrating strong robustness. Nevertheless, middle and deeper transformer layers generally produce better performance than shallow layers, since deeper layers contain richer semantic representations and stronger cross-modal alignment capability. These results indicate that reciprocal multimodal adaptation is particularly effective when applied to semantically expressive transformer stages.

\begin{table}[t]
\centering
\setlength{\tabcolsep}{4pt}
\caption{Ablation Study of VFMs with BRL on RIS Mixed Benchmarks.}
\label{tab_abla_dms}
\resizebox{\linewidth}{!}{
\begin{tabular}{l|ccc|ccc|cc}
\toprule
\multirow{2}{*}{VFMs} & \multicolumn{3}{c|}{RefCOCO} & \multicolumn{3}{c|}{RefCOCO+} & \multicolumn{2}{c}{RefCOCOg} \\ \cmidrule{2-9}
                      & val     & testA   & testB   & val     & testA    & testB   & val(u)     & test(u)    \\
\midrule

\multicolumn{9}{l}{\textit{oIoU}}  \\
\midrule
DINOv3-S+~\cite{dinov3}             & 79.6    & 82.0    & 76.0    & 70.8    & 76.3     & 62.2    & 70.8        & 72.4        \\
DINOv2-B~\cite{dinov2}              & 81.2    & 83.5    & 78.2    & 72.8    & 76.8     & 64.5    & 73.4        & 74.4        \\
DINOv3-B~\cite{dinov3}              & 81.7    & 83.2    & 78.3    & 73.1    & 78.1     & 65.5    & 73.6        & 75.8        \\
EVA2-L~\cite{eva2}                & 82.9    & 84.9    & 79.8    & \textbf{76.0}    & \textbf{80.2}     & \textbf{68.3}    & 75.2        & 77.1        \\
DINOv2-L~\cite{dinov2}              & 82.2    & 84.2    & 79.4    & 74.7    & 78.8     & 66.7    & 74.8        & 76.1        \\
DINOv3-L~\cite{dinov3}              & 82.4    & 84.6    & 79.8    & 74.6    & 79.6     & 67.3    & 75.5        & 76.7        \\
MAE-H~\cite{mae}    & 81.1    & 83.3    & 78.5    & 72.2    & 76.4     & 65.1    & 72.5        & 74.0  \\ 
DINOv2-G~\cite{dinov2}              & 82.6    & 84.5    & 79.3    & 75.1    & 78.8     & 67.8    & \textbf{76.1}        & 75.9            \\
\rowcolor[HTML]{E2F0D9}
DINOv3-H+~\cite{dinov3}             & \textbf{83.5}    & \textbf{85.3}    & \textbf{80.4}    & 75.0    & 80.1     & 68.1    & 75.7        & \textbf{77.4}        \\
\midrule

\multicolumn{9}{l}{\textit{mIoU}}  \\
\midrule
DINOv3-S+~\cite{dinov3}             & 80.7    & 82.4    & 77.6    & 74.0    & 78.3     & 67.1    & 73.3        & 74.3        \\
DINOv2-B~\cite{dinov2}              & 82.2    & 83.9    & 79.7    & 75.9    & 79.4     & 69.7    & 75.5        & 76.3        \\
DINOv3-B~\cite{dinov3}              & 82.4    & 83.8    & 79.4    & 76.3    & 80.4     & 70.2    & 76.0        & 77.3        \\
EVA2-L~\cite{eva2}                & 83.4    & 85.2    & 80.7    & \textbf{78.6}    & 81.9     & \textbf{72.4}    & 77.4        & 78.3        \\
DINOv2-L~\cite{dinov2}              & 83.2    & 84.6    & 80.8    & 77.8    & 81.2     & 71.2    & 76.6        & 77.9        \\
DINOv3-L~\cite{dinov3}              & 83.5    & 85.0    & 81.0    & 77.7    & 81.4     & 71.8    & 77.8        & 78.3        \\
MAE-H~\cite{mae}    & 82.5    & 84.1    & 80.3    & 75.9    & 79.4     & 70.2    & 75.5        & 76.6   \\ 
DINOv2-G~\cite{dinov2}              & 83.7    & 85.0    & 81.4    & 78.3    & 81.4     & 72.4    & 78.0        & 79.0             \\
\rowcolor[HTML]{E2F0D9}
DINOv3-H+~\cite{dinov3}             & \textbf{84.2}    & \textbf{85.6}    & \textbf{81.9}    & 78.0    & \textbf{82.3}     & 72.3    & \textbf{78.0}        & \textbf{79.1}        \\
\bottomrule
\end{tabular}}
\end{table}

\paragraph{Adapter Dimension $d$}

Finally, we study the impact of the adapter hidden dimension $d$, which controls the compression ratio of reciprocal adaptation. As reported in \cref{tab_abla_dim}, increasing $d$ from 16 to 128 consistently improves the segmentation performance, indicating stronger reciprocal representation capacity. In particular, BRL achieves the best overall performance with 77.5\%/80.2\%/74.7\% oIoU and 79.6\%/81.6\%/77.4\% mIoU at $d=128$. However, further increasing $d$ to 256 yields performance degradation while significantly increasing parameters and FLOPs. This suggests that excessively large adapter dimensions may introduce redundancy and weaken parameter efficiency. Considering both performance and efficiency, we adopt $d=64$ as the default setting, which provides a favorable trade-off between reciprocal representation capability and computational overhead.

\begin{figure}[t]
    \centering
    \includegraphics[width=1\linewidth]{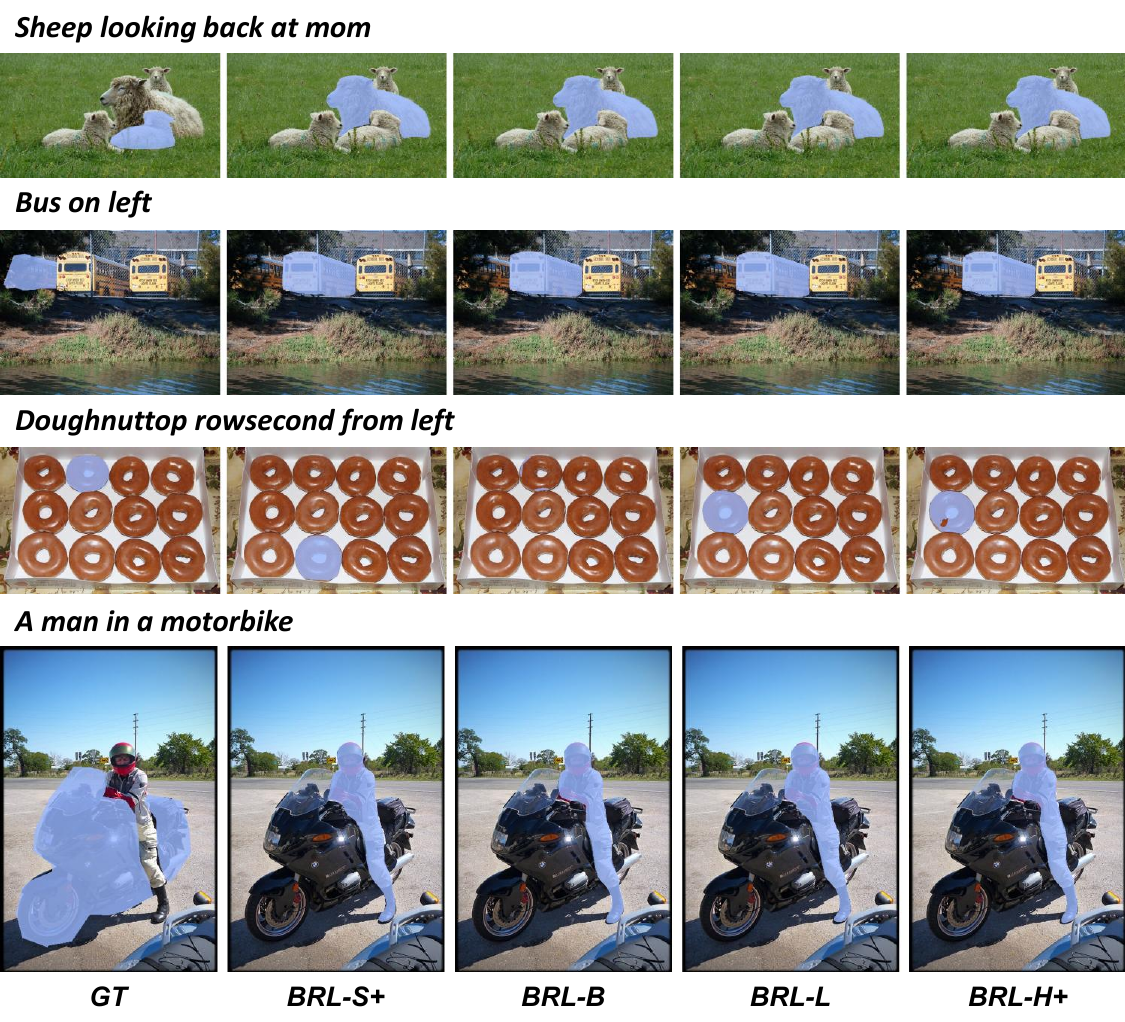}
    \caption{Failure cases of BRL equipped with DINOv3 on RefCOCO.}
    \label{fig_failure}
\end{figure}

\subsubsection{Compatibility with Different VFMs}

To further validate the generalization capability of BRL, we combine our method with various VFMs of different architectures and scales under both the common and the mixed settings. The quantitative results are reported in \cref{tab_abla_vfms} and \cref{tab_abla_dms}, and qualitative comparisons under the mixed setting are illustrated in \cref{fig_mix}.

As shown in \cref{tab_abla_vfms}, BRL consistently improves the performance across different VFMs, including DINO~\cite{dino}, MAE~\cite{mae}, EVA2~\cite{eva2}, DINOv2~\cite{dinov2}, and DINOv3~\cite{dinov3}. Earlier works such as DINO and MAE exhibit relatively limited performance, while more recent large-scale representation learning models achieve substantially better results. In particular, DINOv3 demonstrates the strongest overall performance under almost all scales. For example, DINOv3-B achieves 77.0\%/80.0\%/74.9\% oIoU and 79.3\%/81.4\%/77.1\% mIoU on RefCOCO val/testA/testB, respectively, outperforming DINOv2-B by +1.3\% mIoU testB while using fewer FLOPs. Similar trends are also observed under large-scale settings, where DINOv3-H+ reaches the best overall performance with 80.0\%/82.2\%/76.3\% oIoU and 82.0\%/83.4\%/78.5\% mIoU while only introducing 2.0M parameters (0.2\% of the VFM).

The mixed benchmark results in \cref{tab_abla_dms} further demonstrate the scalability of BRL under larger and more diverse training distributions. As the VFM capacity increases, the segmentation performance improves consistently. Among them, BRL combined with DINOv3-H+ establishes the best overall performance on most RefCOCO and RefCOCOg benchmarks, achieving 83.5\%/85.3\%/80.4\% oIoU and 84.2\%/85.6\%/81.9\% mIoU on RefCOCO val/testA/testB, respectively. Meanwhile, DINOv2-G and EVA2-L also exhibit strong competitiveness on RefCOCOg and RefCOCO+, respectively, indicating that different VFMs may possess different domain preferences.

Overall, these experiments demonstrate that the proposed BRL is highly compatible with diverse VFMs and can consistently transfer the representation advantages of large-scale pretrained visual encoders into referring image segmentation. Moreover, the performance improvements become increasingly significant as the model scale grows, highlighting the strong scalability of BRL in the large foundation model era.

\subsubsection{Failure Cases}

To better understand the limitations of BRL, we further visualize several representative failure cases in \cref{fig_failure}. Generally, most mistakes can be categorized into four types: (1) Complex relational reasoning, where the referring expression contains multi-object interactions, making accurate localization difficult; (2) Severe object occlusion, where the target is heavily covered; (3) Incorrect textual description, where the expression contains inaccurate attributes, resulting in misleading the cross-modal alignment process; (4) Annotation inconsistency, where the provided segmentation mask does not strictly correspond to the actual instance.

Unfortunately, we observe that these challenging samples often lead to consistently incorrect predictions across different model scales. Even when stronger VFMs are adopted, the model still fails rather than showing gradual improvements. This suggests that such errors mainly stem from the intrinsic difficulty of multimodal reasoning and dataset quality issues, instead of insufficient model capacity alone. These observations indicate that future RIS research may require not only stronger foundation models, but also more reliable annotations and enhanced reasoning mechanisms for complex vision-language understanding.

\section{Conclusion}

In this paper, we propose Bidirectional Reciprocal Learning (BRL), a novel parameter-efficient fine-tuning (PEFT) framework for referring image segmentation. Different from existing PEFT methods that mainly focus on adapting visual representations independently, BRL explicitly models bidirectional vision-language interactions through the proposed Reciprocal Attention Adapter (RAA) and Reciprocal Gate Adapter (RGA). Specifically, RAA performs reciprocal cross-modal attention between visual and linguistic features to establish fine-grained semantic correspondence, while RGA further conducts reciprocal global gating to adaptively enhance modality-aware feature responses. By integrating both modules into the transformer architecture through residual adaptation, BRL can effectively transfer multimodal knowledge while maintaining extremely low trainable cost.

Extensive experiments on RefCOCO, RefCOCO+, and RefCOCOg under both common and mixed benchmark settings demonstrate the superiority of the proposed method. BRL consistently outperforms existing PEFT methods by a clear margin on both oIoU and mIoU metrics, and even surpasses recent multimodal large language models despite requiring only a tiny fraction of trainable parameters. Notably, BRL equipped with DINOv3-L already exceeds the performance of some full fine-tuning methods trained under the mixed setting, demonstrating remarkably high adaptation efficiency with substantially lower training data and computational cost. Moreover, BRL exhibits excellent compatibility and scalability across various vision foundation models, including DINO, MAE, EVA2, DINOv2, and DINOv3, and establishes new state-of-the-art performance with large-scale pretrained encoders.

Overall, the proposed BRL provides a scalable and highly efficient solution for multimodal adaptation in referring image segmentation. We hope this work can offer useful insights for future research on parameter-efficient multimodal learning and encourage further exploration of efficient adaptation strategies for large vision-language foundation models.

\bibliography{IEEEabrv,ref}

@STRING{CVPR      = "Proc. IEEE Conf. Comput. Vis. Pattern Recognit. (CVPR)"}

@STRING{ICCV      = "Proc. IEEE/CVF Int. Conf. Comput. Vis. (ICCV)"}

@STRING{ECCV      = "Proc. of Eur. Conf. on Comput. Vis. (ECCV)"}

@STRING{NIPS      = "Proc. Adv. Neural Inf. Process. Syst. (NeurIPS)"}

@STRING{ICLR      = "Proc. Int. Conf. Learn. Represent. (ICLR)"}

@STRING{ICML      = "Proc. Int. Conf. Mach. Learn. (ICML)"}

@STRING{NAACL      = "NAACL"}

@STRING{EMNLP      = "EMNLP"}

@STRING{AAAI      = "AAAI Conf. Artif. Intell. (AAAI)"}

@STRING{ACMMM     = "ACM Int. Conf. Multimedia. (ACM MM)"}

@STRING{IJCV       = "Int. J. Comput. Vis."}

@STRING{TPAMI       = "{IEEE} Trans. Pattern Anal. Mach. Intell."}

@STRING{TGRS        = "{IEEE} Trans. Geosci. Remote Sens."}

@STRING{GRSM   = "{IEEE} Geosci. Remote Sens. Mag."}

@STRING{TCSVT      = "{IEEE} Trans. Circuits Syst. Video Technol."}

@STRING{TIP         = "{IEEE} Trans. Image Process."}

@STRING{TMM         = "{IEEE} Trans. Multimedia."}

@STRING{TMI         = "{IEEE} Trans. Med. Imag."}

@STRING{TMLR      = "Trans. Mach. Learn. Res."}

@STRING{IS      = "Inf. Sci."}

@STRING{MIA      = "Med. Image Anal."}

@inproceedings{vit,
  title={An Image is Worth 16x16 Words: Transformers for Image Recognition at Scale},
  author={Dosovitskiy, Alexey and Beyer, Lucas and Kolesnikov, Alexander and Weissenborn, Dirk and Zhai, Xiaohua and Unterthiner, Thomas and Dehghani, Mostafa and Minderer, Matthias and Heigold, Georg and Gelly, Sylvain and others},
  booktitle=ICLR,
  pages={1--22},
  year={2021}
}

@inproceedings{swin,
  title={Swin transformer: Hierarchical vision transformer using shifted windows},
  author={Liu, Ze and Lin, Yutong and Cao, Yue and Hu, Han and Wei, Yixuan and Zhang, Zheng and Lin, Stephen and Guo, Baining},
  booktitle=ICCV,
  pages={10012--10022},
  year={2021}
}

@article{vmamba,
  title={VMamba: Visual State Space Model},
  author={Liu, Yue and Tian, Yunjie and Zhao, Yuzhong and Yu, Hongtian and Xie, Lingxi and Wang, Yaowei and Ye, Qixiang and Liu, Yunfan},
  journal=NIPS,
  pages={103031--103063},
  year={2024}
}

@inproceedings{bert,
  title={Bert: Pre-training of deep bidirectional transformers for language understanding},
  author={Devlin, Jacob and Chang, Ming-Wei and Lee, Kenton and Toutanova, Kristina},
  booktitle=NAACL,
  pages={4171--4186},
  year={2019}
}

@article{roberta,
  title={Roberta: A robustly optimized bert pretraining approach},
  author={Liu, Yinhan and Ott, Myle and Goyal, Naman and Du, Jingfei and Joshi, Mandar and Chen, Danqi and Levy, Omer and Lewis, Mike and Zettlemoyer, Luke and Stoyanov, Veselin},
  journal={arXiv preprint arXiv:1907.11692},
  year={2019}
}

@inproceedings{clip,
  title={Learning transferable visual models from natural language supervision},
  author={Radford, Alec and Kim, Jong Wook and Hallacy, Chris and Ramesh, Aditya and Goh, Gabriel and Agarwal, Sandhini and Sastry, Girish and Askell, Amanda and Mishkin, Pamela and Clark, Jack and others},
  booktitle=ICML,
  pages={8748--8763},
  year={2021},
  organization={PMLR}
}

@article{eva_clip,
  title={Eva-clip: Improved training techniques for clip at scale},
  author={Sun, Quan and Fang, Yuxin and Wu, Ledell and Wang, Xinlong and Cao, Yue},
  journal={arXiv preprint arXiv:2303.15389},
  year={2023}
}

@inproceedings{sam,
  title={Segment anything},
  author={Kirillov, Alexander and Mintun, Eric and Ravi, Nikhila and Mao, Hanzi and Rolland, Chloe and Gustafson, Laura and Xiao, Tete and Whitehead, Spencer and Berg, Alexander C and Lo, Wan-Yen and others},
  booktitle=ICCV,
  pages={4015--4026},
  year={2023}
}

@article{llava,
  title={Visual instruction tuning},
  author={Liu, Haotian and Li, Chunyuan and Wu, Qingyang and Lee, Yong Jae},
  journal=NIPS,
  volume={36},
  pages={34892--34916},
  year={2023}
}

@article{dinov2,
  title={DINOv2: Learning Robust Visual Features without Supervision},
  author={Oquab, Maxime and Darcet, Timoth{\'e}e and Moutakanni, Th{\'e}o and Vo, Huy and Szafraniec, Marc and Khalidov, Vasil and Fernandez, Pierre and Haziza, Daniel and Massa, Francisco and El-Nouby, Alaaeldin and others},
  journal=TMLR,
  pages={1--31},
  year={2024}
}

@article{dinov3,
  title={Dinov3},
  author={Sim{\'e}oni, Oriane and Vo, Huy V and Seitzer, Maximilian and Baldassarre, Federico and Oquab, Maxime and Jose, Cijo and Khalidov, Vasil and Szafraniec, Marc and Yi, Seungeun and Ramamonjisoa, Micha{\"e}l and others},
  journal={arXiv preprint arXiv:2508.10104},
  year={2025}
}

@inproceedings{mattnet,
  title={Mattnet: Modular attention network for referring expression comprehension},
  author={Yu, Licheng and Lin, Zhe and Shen, Xiaohui and Yang, Jimei and Lu, Xin and Bansal, Mohit and Berg, Tamara L},
  booktitle=CVPR,
  pages={1307--1315},
  year={2018}
}

@inproceedings{cmatt,
  title={Improving referring expression grounding with cross-modal attention-guided erasing},
  author={Liu, Xihui and Wang, Zihao and Shao, Jing and Wang, Xiaogang and Li, Hongsheng},
  booktitle=CVPR,
  pages={1950--1959},
  year={2019}
}

@inproceedings{efn,
  title={Encoder fusion network with co-attention embedding for referring image segmentation},
  author={Feng, Guang and Hu, Zhiwei and Zhang, Lihe and Lu, Huchuan},
  booktitle=CVPR,
  pages={15506--15515},
  year={2021}
}

@inproceedings{cris,
  title={Cris: Clip-driven referring image segmentation},
  author={Wang, Zhaoqing and Lu, Yu and Li, Qiang and Tao, Xunqiang and Guo, Yandong and Gong, Mingming and Liu, Tongliang},
  booktitle=CVPR,
  pages={11686--11695},
  year={2022}
}

@inproceedings{restr,
  title={Restr: Convolution-free referring image segmentation using transformers},
  author={Kim, Namyup and Kim, Dongwon and Lan, Cuiling and Zeng, Wenjun and Kwak, Suha},
  booktitle=CVPR,
  pages={18145--18154},
  year={2022}
}

@inproceedings{lavt,
  title={Lavt: Language-aware vision transformer for referring image segmentation},
  author={Yang, Zhao and Wang, Jiaqi and Tang, Yansong and Chen, Kai and Zhao, Hengshuang and Torr, Philip HS},
  booktitle=CVPR,
  pages={18155--18165},
  year={2022}
}

@article{vlt,
  title={VLT: Vision-language transformer and query generation for referring segmentation},
  author={Ding, Henghui and Liu, Chang and Wang, Suchen and Jiang, Xudong},
  journal=TPAMI,
  volume={45},
  number={6},
  pages={7900--7916},
  year={2022},
  publisher={IEEE}
}

@inproceedings{cgformer,
  title={Contrastive grouping with transformer for referring image segmentation},
  author={Tang, Jiajin and Zheng, Ge and Shi, Cheng and Yang, Sibei},
  booktitle=CVPR,
  pages={23570--23580},
  year={2023}
}

@inproceedings{rela,
  title={Gres: Generalized referring expression segmentation},
  author={Liu, Chang and Ding, Henghui and Jiang, Xudong},
  booktitle=CVPR,
  pages={23592--23601},
  year={2023}
}

@inproceedings{polyformer,
  title={Polyformer: Referring image segmentation as sequential polygon generation},
  author={Liu, Jiang and Ding, Hui and Cai, Zhaowei and Zhang, Yuting and Satzoda, Ravi Kumar and Mahadevan, Vijay and Manmatha, R},
  booktitle=CVPR,
  pages={18653--18663},
  year={2023}
}

@inproceedings{vglaw,
  title={Language adaptive weight generation for multi-task visual grounding},
  author={Su, Wei and Miao, Peihan and Dou, Huanzhang and Wang, Gaoang and Qiao, Liang and Li, Zheyang and Li, Xi},
  booktitle=CVPR,
  pages={10857--10866},
  year={2023}
}

@inproceedings{dmmi,
  title={Beyond one-to-one: Rethinking the referring image segmentation},
  author={Hu, Yutao and Wang, Qixiong and Shao, Wenqi and Xie, Enze and Li, Zhenguo and Han, Jungong and Luo, Ping},
  booktitle=ICCV,
  pages={4067--4077},
  year={2023}
}

@inproceedings{etris,
  title={Bridging vision and language encoders: Parameter-efficient tuning for referring image segmentation},
  author={Xu, Zunnan and Chen, Zhihong and Zhang, Yong and Song, Yibing and Wan, Xiang and Li, Guanbin},
  booktitle=ICCV,
  pages={17503--17512},
  year={2023}
}

@article{efn_tpami,
  author={Feng, Guang and Zhang, Lihe and Sun, Jiayu and Hu, Zhiwei and Lu, Huchuan},
  journal=TPAMI, 
  title={Referring Segmentation via Encoder-Fused Cross-Modal Attention Network}, 
  year={2023},
  volume={45},
  number={6},
  pages={7654-7667},
}

@article{refsegformer,
  title={Toward Robust Referring Image Segmentation},
  author={Wu, Jianzong and Li, Xiangtai and Li, Xia and Ding, Henghui and Tong, Yunhai and Tao, Dacheng},
  journal=TIP,
  year={2024},
  publisher={IEEE}
}

@inproceedings{barleria,
  title={Barleria: An efficient tuning framework for referring image segmentation},
  author={Wang, Yaoming and Li, Jin and Zhang, Xiaopeng and Shi, Bowen and Li, Chenglin and Dai, Wenrui and Xiong, Hongkai and Tian, Qi},
  booktitle=ICLR,
  pages={1--12},
  year={2024}
}

@inproceedings{magnet,
  title={Mask grounding for referring image segmentation},
  author={Chng, Yong Xien and Zheng, Henry and Han, Yizeng and Qiu, Xuchong and Huang, Gao},
  booktitle=CVPR,
  pages={26573--26583},
  year={2024}
}

@inproceedings{lqmformer,
  title={Lqmformer: Language-aware query mask transformer for referring image segmentation},
  author={Shah, Nisarg A and VS, Vibashan and Patel, Vishal M},
  booktitle=CVPR,
  pages={12903--12913},
  year={2024}
}

@inproceedings{remamber,
  title={Remamber: Referring image segmentation with mamba twister},
  author={Yang, Yuhuan and Ma, Chaofan and Yao, Jiangchao and Zhong, Zhun and Zhang, Ya and Wang, Yanfeng},
  booktitle=ECCV,
  pages={108--126},
  year={2024},
  organization={Springer}
}

@inproceedings{asda,
  title={Adaptive selection based referring image segmentation},
  author={Yue, Pengfei and Lin, Jianghang and Zhang, Shengchuan and Hu, Jie and Lu, Yilin and Niu, Hongwei and Ding, Haixin and Zhang, Yan and Jiang, Guannan and Cao, Liujuan and others},
  booktitle=ACMMM,
  pages={1101--1110},
  year={2024}
}

@inproceedings{risclip,
  title={Extending CLIPs Image-Text Alignment to Referring Image Segmentation},
  author={Kim, Seoyeon and Kang, Minguk and Kim, Dongwon and Park, Jaesik and Kwak, Suha},
  booktitle=NAACL,
  volume={1},
  pages={4611--4628},
  year={2024},
  organization={ACL}
}

@article{lavt_tpami,
  author={Yang, Zhao and Wang, Jiaqi and Ye, Xubing and Tang, Yansong and Chen, Kai and Zhao, Hengshuang and Torr, Philip H. S.},
  journal=TPAMI, 
  title={Language-Aware Vision Transformer for Referring Segmentation}, 
  year={2025},
  volume={47},
  number={7},
  pages={5238-5255},
}

@inproceedings{detris,
  title={Densely connected parameter-efficient tuning for referring image segmentation},
  author={Huang, Jiaqi and Xu, Zunnan and Liu, Ting and Liu, Yong and Han, Haonan and Yuan, Kehong and Li, Xiu},
  booktitle=AAAI,
  volume={39},
  number={4},
  pages={3653--3661},
  year={2025}
}

@ARTICLE{cmirnet,
  author={Xu, Mingzhu and Xiao, Tianxiang and Liu, Yutong and Tang, Haoyu and Hu, Yupeng and Nie, Liqiang},
  journal=TCSVT, 
  title={CMIRNet: Cross-Modal Interactive Reasoning Network for Referring Image Segmentation}, 
  year={2025},
  volume={35},
  number={4},
  pages={3234-3249}
}

@ARTICLE{cmanet,
  author={Pan, Xiong and Xie, Xuemei and Yang, Jianxiu and Song, Xiaodan and Shi, Guangming},
  journal=TMM, 
  title={CMANet: Context-aware Mutual Attention Network for Referring Image Segmentation}, 
  year={2025},
  volume={},
  number={},
  pages={1-12}
}

@ARTICLE{soc,
  author={Liu, Yong and Luo, Zhuoyan and Xiao, Yicheng and Wang, Yitong and Li, Shuyan and Li, Xiu and Yang, Yujiu and Tang, Yansong},
  journal=TPAMI, 
  title={Semantic-Assisted Object Clustering for Multi-Modal Referring Video Segmentation}, 
  year={2026},
  volume={48},
  number={1},
  pages={572-590},
}

@inproceedings{lisa,
  title={Lisa: Reasoning segmentation via large language model},
  author={Lai, Xin and Tian, Zhuotao and Chen, Yukang and Li, Yanwei and Yuan, Yuhui and Liu, Shu and Jia, Jiaya},
  booktitle=CVPR,
  pages={9579--9589},
  year={2024}
}

@inproceedings{gsva,
  title={Gsva: Generalized segmentation via multimodal large language models},
  author={Xia, Zhuofan and Han, Dongchen and Han, Yizeng and Pan, Xuran and Song, Shiji and Huang, Gao},
  booktitle=CVPR,
  pages={3858--3869},
  year={2024}
}

@inproceedings{M2SA,
  title={MMR: A Large-scale Benchmark Dataset for Multi-target and Multi-granularity Reasoning Segmentation},
  author={Jang, Donggon and Cho, Yucheol and Lee, Suin and Kim, Taehyeon and Kim, Daeshik},
  booktitle=ICLR,
  pages={1--15},
  year={2025}
}

@inproceedings{segllm,
  title={Segllm: Multi-round reasoning segmentation},
  author={Wang, XuDong and Zhang, Shaolun and Li, Shufan and Kallidromitis, Konstantinos and Li, Kehan and Kato, Yusuke and Kozuka, Kazuki and Darrell, Trevor},
  booktitle=ICLR,
  pages={1--14},
  year={2025}
}

@inproceedings{hiea2g,
  title={Hierarchical alignment-enhanced adaptive grounding network for generalized referring expression comprehension},
  author={Wang, Yaxian and Ding, Henghui and He, Shuting and Jiang, Xudong and Wei, Bifan and Liu, Jun},
  booktitle=AAAI,
  volume={39},
  number={8},
  pages={8042--8050},
  year={2025}
}

@inproceedings{read,
  title={Reasoning to attend: Try to understand how< seg> token works},
  author={Qian, Rui and Yin, Xin and Dou, Dejing},
  booktitle=CVPR,
  pages={24722--24731},
  year={2025}
}

@inproceedings{popen,
  title={Popen: Preference-based optimization and ensemble for lvlm-based reasoning segmentation},
  author={Zhu, Lanyun and Chen, Tianrun and Xu, Qianxiong and Liu, Xuanyi and Ji, Deyi and Wu, Haiyang and Soh, De Wen and Liu, Jun},
  booktitle=CVPR,
  pages={30231--30240},
  year={2025}
}

@article{lscf,
  title={LSCF: Long-term Semantic-guidance ConvFormer for Referring Remote Sensing Image Segmentation},
  author={Ma, Qin and Li, Lingling and Lu, Xiaoqiang and Jiao, Licheng and Liu, Fang and Ma, Wenping and Liu, Xu and Sun, Long},
  journal=TGRS,
  year={2025},
  publisher={IEEE}
}

@ARTICLE{rrsecs,
  author={Lu, Xiaoqiang and Sun, Long and Li, Lingling and Jiao, Licheng and Yang, Yuting and Huang, Zhongjian and Chai, Jinming and Liu, Xu and Liu, Fang and Ma, Wenping and Yang, Shuyuan},
  journal=GRSM, 
  title={RRSECS: Referring remote sensing expression comprehension and segmentation}, 
  year={2025},
  volume={13},
  number={3},
  pages={440-467}
}

@ARTICLE{seeformer,
  author={Chai, Jinming and Jiao, Licheng and Lu, Xiaoqiang and Li, Lingling and Liu, Fang and Sun, Long and Liu, Xu and Ma, Wenping and Li, Weibin},
  journal=TPAMI, 
  title={Like Human Rethinking: Contour Transformer AutoRegression for Referring Remote Sensing Interpretation}, 
  year={2026},
  volume={},
  number={},
  pages={1-18},
}

@article{lvit,
  title={Lvit: language meets vision transformer in medical image segmentation},
  author={Li, Zihan and Li, Yunxiang and Li, Qingde and Wang, Puyang and Guo, Dazhou and Lu, Le and Jin, Dakai and Zhang, You and Hong, Qingqi},
  journal=TMI,
  year={2023},
  volume={43},
  number={1},
  pages={96-107},
  publisher={IEEE}
}

@article{reclmis,
  title={Cross-modal conditioned reconstruction for language-guided medical image segmentation},
  author={Huang, Xiaoshuang and Li, Hongxiang and Cao, Meng and Chen, Long and You, Chenyu and An, Dong},
  journal=TMI,
  year={2024},
  volume={44},
  number={4},
  pages={1821-1835},
  publisher={IEEE}
}

@article{tvenet,
  title={Driven by textual knowledge: A Text-View Enhanced Knowledge Transfer Network for lung infection region segmentation},
  author={Fang, Lexin and Li, Xuemei and Xu, Yunyang and Zhang, Fan and Zhang, Caiming},
  journal=MIA,
  pages={103625},
  year={2025},
  publisher={Elsevier}
}

@inproceedings{refcoco,
  title={Modeling context in referring expressions},
  author={Yu, Licheng and Poirson, Patrick and Yang, Shan and Berg, Alexander C and Berg, Tamara L},
  booktitle=ECCV,
  pages={69--85},
  year={2016}
}

@inproceedings{gref1,
  title={Generation and comprehension of unambiguous object descriptions},
  author={Mao, Junhua and Huang, Jonathan and Toshev, Alexander and Camburu, Oana and Yuille, Alan L and Murphy, Kevin},
  booktitle=CVPR,
  pages={11--20},
  year={2016}
}

@inproceedings{gref2,
  title={Modeling context between objects for referring expression understanding},
  author={Nagaraja, Varun K and Morariu, Vlad I and Davis, Larry S},
  booktitle=ECCV,
  pages={792--807},
  year={2016},
  organization={Springer}
}

@article{peft_llm,
  title={Parameter-Efficient Fine-Tuning for Large Models: A Comprehensive Survey},
  author={Han, Zeyu and Gao, Chao and Liu, Jinyang and Zhang, Jeff and Zhang, Sai Qian},
  journal=TMLR,
  year={2024},
}

@article{peft_vfm,
  title={Parameter-efficient fine-tuning for pre-trained vision models: A survey},
  author={Xin, Yi and Luo, Siqi and Zhou, Haodi and Du, Junlong and Liu, Xiaohong and Fan, Yue and Li, Qing and Du, Yuntao},
  journal={arXiv preprint arXiv:2402.02242},
  year={2024}
}

@article{peft_fm,
  title={Parameter-efficient fine-tuning for foundation models},
  author={Zhang, Dan and Feng, Tao and Xue, Lilong and Wang, Yuandong and Dong, Yuxiao and Tang, Jie},
  journal={arXiv preprint arXiv:2501.13787},
  year={2025}
}

@inproceedings{lora,
  title={Lora: Low-rank adaptation of large language models.},
  author={Hu, Edward J and Shen, Yelong and Wallis, Phillip and Allen-Zhu, Zeyuan and Li, Yuanzhi and Wang, Shean and Wang, Lu and Chen, Weizhu and others},
  booktitle=ICLR,
  pages={1--12},
  year={2022}
}

@article{lora_fa,
  title={Lora-fa: Memory-efficient low-rank adaptation for large language models fine-tuning},
  author={Zhang, Longteng and Zhang, Lin and Shi, Shaohuai and Chu, Xiaowen and Li, Bo},
  journal={arXiv preprint arXiv:2308.03303},
  year={2023}
}

@inproceedings{lorand,
  title={1\% vs 100\%: Parameter-efficient low rank adapter for dense predictions},
  author={Yin, Dongshuo and Yang, Yiran and Wang, Zhechao and Yu, Hongfeng and Wei, Kaiwen and Sun, Xian},
  booktitle=CVPR,
  pages={20116--20126},
  year={2023}
}

@article{qlora,
  title={Qlora: Efficient finetuning of quantized llms},
  author={Dettmers, Tim and Pagnoni, Artidoro and Holtzman, Ari and Zettlemoyer, Luke},
  journal=NIPS,
  volume={36},
  pages={10088--10115},
  year={2023}
}

@inproceedings{prompt_tuning,
  title={The power of scale for parameter-efficient prompt tuning},
  author={Lester, Brian and Al-Rfou, Rami and Constant, Noah},
  booktitle=EMNLP,
  pages={3045--3059},
  year={2021}
}

@inproceedings{vpt,
  title={Visual prompt tuning},
  author={Jia, Menglin and Tang, Luming and Chen, Bor-Chun and Cardie, Claire and Belongie, Serge and Hariharan, Bharath and Lim, Ser-Nam},
  booktitle=ECCV,
  pages={709--727},
  year={2022},
  organization={Springer}
}

@inproceedings{qformer,
  title={Blip-2: Bootstrapping language-image pre-training with frozen image encoders and large language models},
  author={Li, Junnan and Li, Dongxu and Savarese, Silvio and Hoi, Steven},
  booktitle=ICML,
  pages={19730--19742},
  year={2023},
  organization={PMLR}
}

@inproceedings{adapter,
  title={Parameter-efficient transfer learning for NLP},
  author={Houlsby, Neil and Giurgiu, Andrei and Jastrzebski, Stanislaw and Morrone, Bruna and De Laroussilhe, Quentin and Gesmundo, Andrea and Attariyan, Mona and Gelly, Sylvain},
  booktitle=ICML,
  pages={2790--2799},
  year={2019},
  organization={PMLR}
}

@article{adaptformer,
  title={Adaptformer: Adapting vision transformers for scalable visual recognition},
  author={Chen, Shoufa and Ge, Chongjian and Tong, Zhan and Wang, Jiangliu and Song, Yibing and Wang, Jue and Luo, Ping},
  journal=NIPS,
  volume={35},
  pages={16664--16678},
  year={2022}
}

@inproceedings{vitadapter,
  title={Vision Transformer Adapter for Dense Predictions},
  author={Chen, Zhe and Duan, Yuchen and Wang, Wenhai and He, Junjun and Lu, Tong and Dai, Jifeng and Qiao, Yu},
  booktitle=ICLR,
  pages={1--14},
  year={2023}
}

@article{clipadapter,
  title={Clip-adapter: Better vision-language models with feature adapters},
  author={Gao, Peng and Geng, Shijie and Zhang, Renrui and Ma, Teli and Fang, Rongyao and Zhang, Yongfeng and Li, Hongsheng and Qiao, Yu},
  journal=IJCV,
  volume={132},
  number={2},
  pages={581--595},
  year={2024},
  publisher={Springer}
}

@inproceedings{mscoco,
  title={Microsoft coco: Common objects in context},
  author={Lin, Tsung-Yi and Maire, Michael and Belongie, Serge and Hays, James and Perona, Pietro and Ramanan, Deva and Doll{\'a}r, Piotr and Zitnick, C Lawrence},
  booktitle=ECCV,
  pages={740--755},
  year={2014},
  organization={Springer}
}

@inproceedings{vitdet,
  title={Exploring plain vision transformer backbones for object detection},
  author={Li, Yanghao and Mao, Hanzi and Girshick, Ross and He, Kaiming},
  booktitle=ECCV,
  pages={280--296},
  year={2022},
  organization={Springer}
}

@inproceedings{deeplabv3+,
  title={Encoder-decoder with atrous separable convolution for semantic image segmentation},
  author={Chen, Liang-Chieh and Zhu, Yukun and Papandreou, George and Schroff, Florian and Adam, Hartwig},
  booktitle=ECCV,
  pages={801--818},
  year={2018}
}

@inproceedings{upernet,
  title={Unified perceptual parsing for scene understanding},
  author={Xiao, Tete and Liu, Yingcheng and Zhou, Bolei and Jiang, Yuning and Sun, Jian},
  booktitle=ECCV,
  pages={418--434},
  year={2018}
}

@article{segformer,
  title={SegFormer: Simple and efficient design for semantic segmentation with transformers},
  author={Xie, Enze and Wang, Wenhai and Yu, Zhiding and Anandkumar, Anima and Alvarez, Jose M and Luo, Ping},
  journal=NIPS,
  volume={34},
  pages={12077--12090},
  year={2021}
}

@inproceedings{mask2former,
  title={Masked-attention mask transformer for universal image segmentation},
  author={Cheng, Bowen and Misra, Ishan and Schwing, Alexander G and Kirillov, Alexander and Girdhar, Rohit},
  booktitle=CVPR,
  pages={1290--1299},
  year={2022}
}

@article{lsst,
  title={Simple and efficient: A semisupervised learning framework for remote sensing image semantic segmentation},
  author={Lu, Xiaoqiang and Jiao, Licheng and Liu, Fang and Yang, Shuyuan and Liu, Xu and Feng, Zhixi and Li, Lingling and Chen, Puhua},
  journal=TGRS,
  volume={60},
  pages={1--16},
  year={2022},
  publisher={IEEE}
}

@article{wscl,
  title={Weak-to-strong consistency learning for semisupervised image segmentation},
  author={Lu, Xiaoqiang and Jiao, Licheng and Li, Lingling and Liu, Fang and Liu, Xu and Yang, Shuyuan and Feng, Zhixi and Chen, Puhua},
  journal=TGRS,
  volume={61},
  pages={1--15},
  year={2023},
  publisher={IEEE}
}

@article{speed,
  title={Self Pseudo Entropy Knowledge Distillation for Semi-supervised Semantic Segmentation},
  author={Lu, Xiaoqiang and Jiao, Licheng and Li, Lingling and Liu, Fang and Liu, Xu and Yang, Shuyuan},
  journal=TCSVT,
  volume={34},
  pages={7359--7372},
  year={2024},
  publisher={IEEE}
}

@article{umcl,
  title={Uncertainty-aware Semi-supervised Learning Segmentation for Remote Sensing Images},
  author={Lu, Xiaoqiang and Li, Lingling and Jiao, Licheng and Liu, Xu and Liu, Fang and Ma, Wenping and Yang, Shuyuan},
  journal=TMM,
  year={2025},
  volume={27},
  pages={5548-5562},
  publisher={IEEE}
}

@inproceedings{dino,
  title={Emerging properties in self-supervised vision transformers},
  author={Caron, Mathilde and Touvron, Hugo and Misra, Ishan and J{\'e}gou, Herv{\'e} and Mairal, Julien and Bojanowski, Piotr and Joulin, Armand},
  booktitle=ICCV,
  pages={9650--9660},
  year={2021}
}

@inproceedings{mae,
  title={Masked autoencoders are scalable vision learners},
  author={He, Kaiming and Chen, Xinlei and Xie, Saining and Li, Yanghao and Doll{\'a}r, Piotr and Girshick, Ross},
  booktitle=CVPR,
  pages={16000--16009},
  year={2022}
}

@article{eva2,
  title={Eva-02: A visual representation for neon genesis},
  author={Fang, Yuxin and Sun, Quan and Wang, Xinggang and Huang, Tiejun and Wang, Xinlong and Cao, Yue},
  journal={Image and Vision Computing},
  volume={149},
  pages={105171},
  year={2024},
  publisher={Elsevier}
}

@article{cmpc_tpami,
  author={Liu, Si and Hui, Tianrui and Huang, Shaofei and Wei, Yunchao and Li, Bo and Li, Guanbin},
  journal=TPAMI, 
  title={Cross-Modal Progressive Comprehension for Referring Segmentation}, 
  year={2022},
  volume={44},
  number={9},
  pages={4761-4775}
}
\bibliographystyle{IEEEtran}

\begin{IEEEbiography}[{\includegraphics[width=1in,height=1.25in,clip,keepaspectratio]{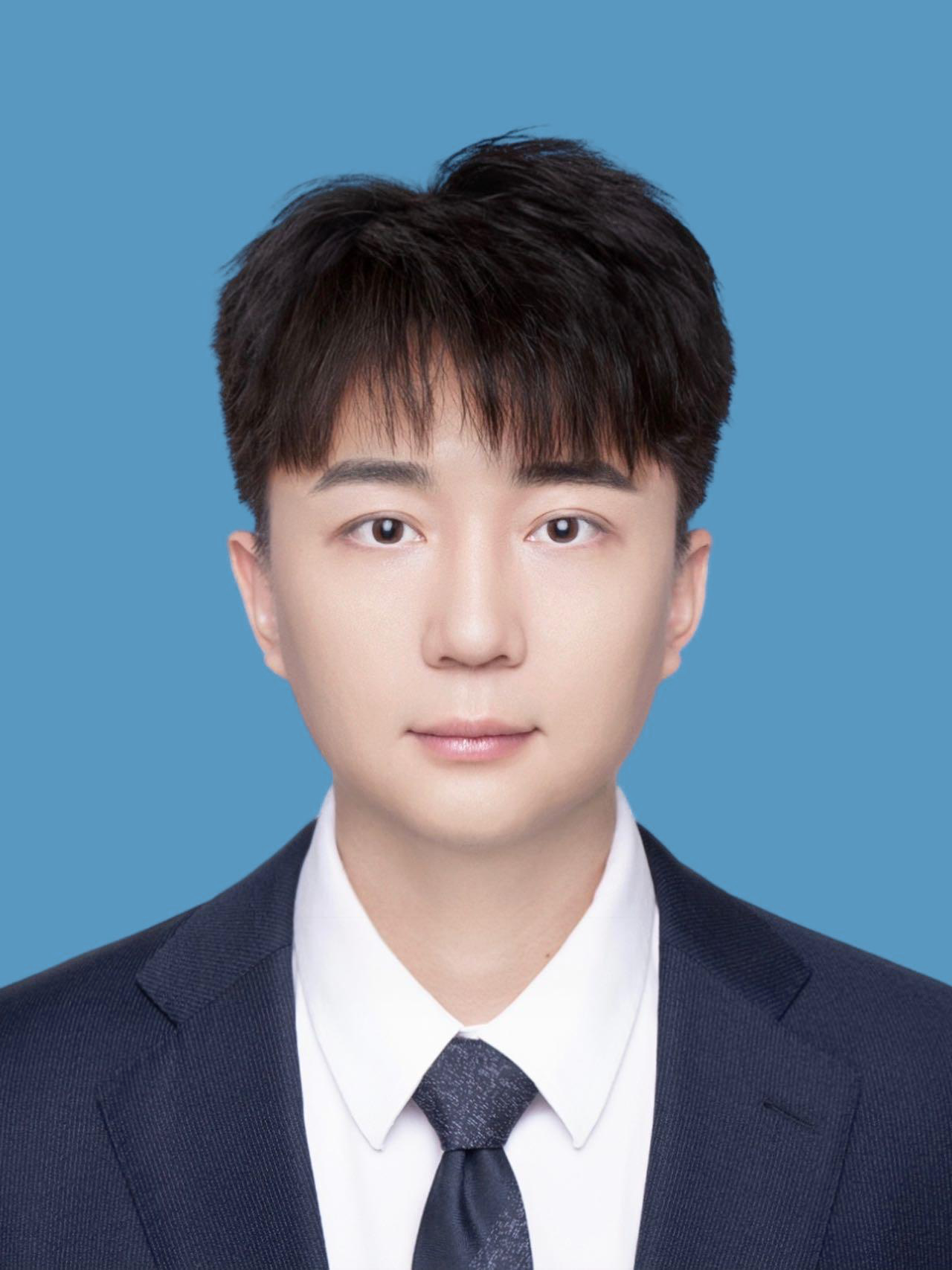}}]{Xiaoqiang Lu} (Member, IEEE) received the B.S. degree in Information Countermeasure Technique from Xidian University, Xi’an, China, in 2020, and the Ph.D. degree in Computer Science and Technology from Xidian University, Xi’an, China, in 2025. He is currently a Postdoctoral Researcher with the School of Artificial Intelligence at Xidian University. 

His research interests include remote sensing interpretation, computer vision, multi-modal learning, and foundation model.
\end{IEEEbiography}


\begin{IEEEbiography}[{\includegraphics[width=1in,height=1.25in,clip,keepaspectratio]{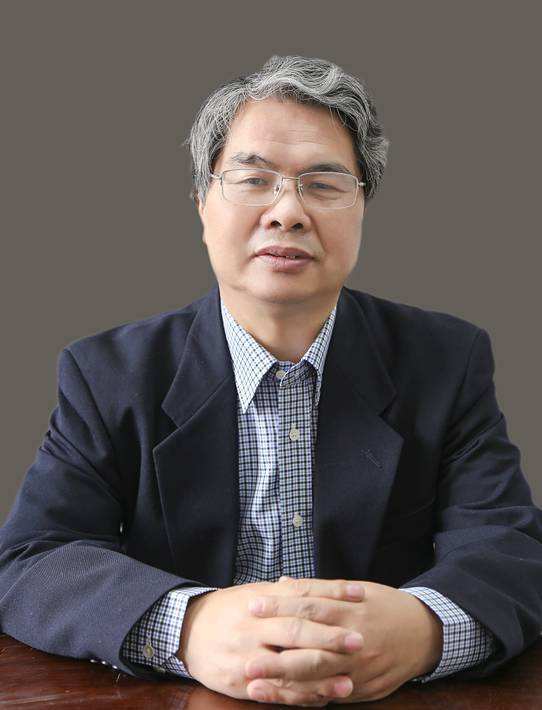}}]{Licheng Jiao}
(Life Fellow, IEEE) received the B.S. degree from Shanghai Jiaotong University, Shanghai, China, in 1982, and the M.S. and Ph.D. degrees from Xi’an Jiaotong University, Xi’an, China, in 1984 and 1990, respectively. Since 1992, he has been a Professor at the School of Electronic Engineering, Xidian University, Xi’an, where he is currently the Director of the Key Laboratory of Intelligent Perception and Image Understanding of the Ministry of Education of China. He is in charge of about 40 important scientific research projects. He has authored/coauthored more than 20 monographs and 100 papers in international journals and conferences. His research interests include image processing, natural computation, machine learning, and intelligent information processing.

Dr. Jiao is a Foreign Member of the Academia Europaea and the Russian Academy of Natural Sciences, a fellow of IET/CAAI/CIE/CCF/CAA, a Councilor of the Chinese Institute of Electronics, a committee member of the Chinese Committee of Neural Networks, and an Expert of the Academic Degrees Committee of the State Council.
\end{IEEEbiography}


\begin{IEEEbiography}[{\includegraphics[width=1in,height=1.25in,clip,keepaspectratio]{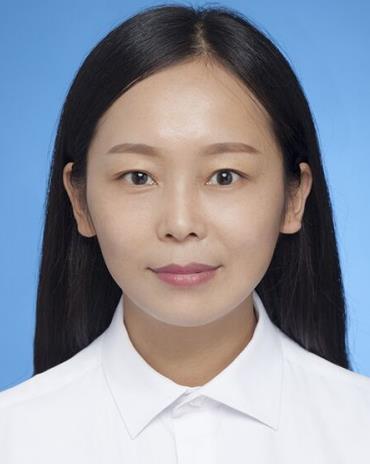}}]{Lingling Li}
(Senior Member, IEEE) received the B.S. and Ph.D. degrees from Xidian University, Xi’an, China, in 2011 and 2017, respectively.

From 2013 to 2014, she was an Exchange Ph.D. Student with the Intelligent Systems Group, Department of Computer Science and Artificial Intelligence, University of the Basque Country UPV/EHU, Leioa, Spain. She is currently an Associate Professor with the School of Artificial Intelligence, Xidian University. Her research interests include quantum evolutionary optimization and deep learning.
\end{IEEEbiography}


\begin{IEEEbiography}[{\includegraphics[width=1in,height=1.25in,clip,keepaspectratio]{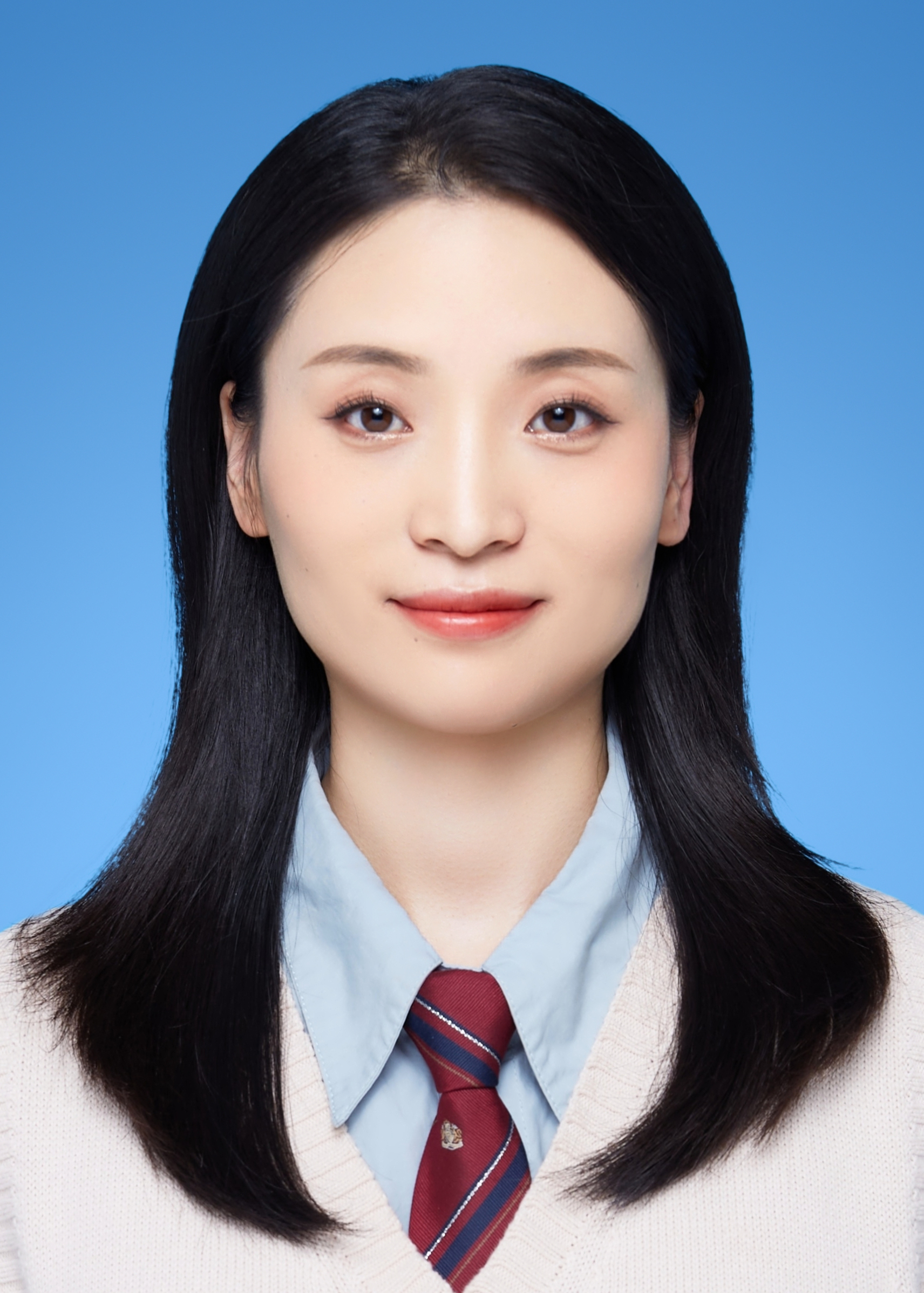}}]{Yuting Yang} 
(Member, IEEE) received the B.S. degree from Northwest University, Xi'an,
China, in 2018, and the PhD degree from Xidian University, Xi'an, China, in 2024. She is currently a Postdoctoral Researcher with the School of Artificial Intelligence at Xidian University. 

Her research interests include computer vision, the interpretability of deep learning, and multiscale geometric analysis.
\end{IEEEbiography}


\begin{IEEEbiography}[{\includegraphics[width=1in,height=1.25in,clip,keepaspectratio]{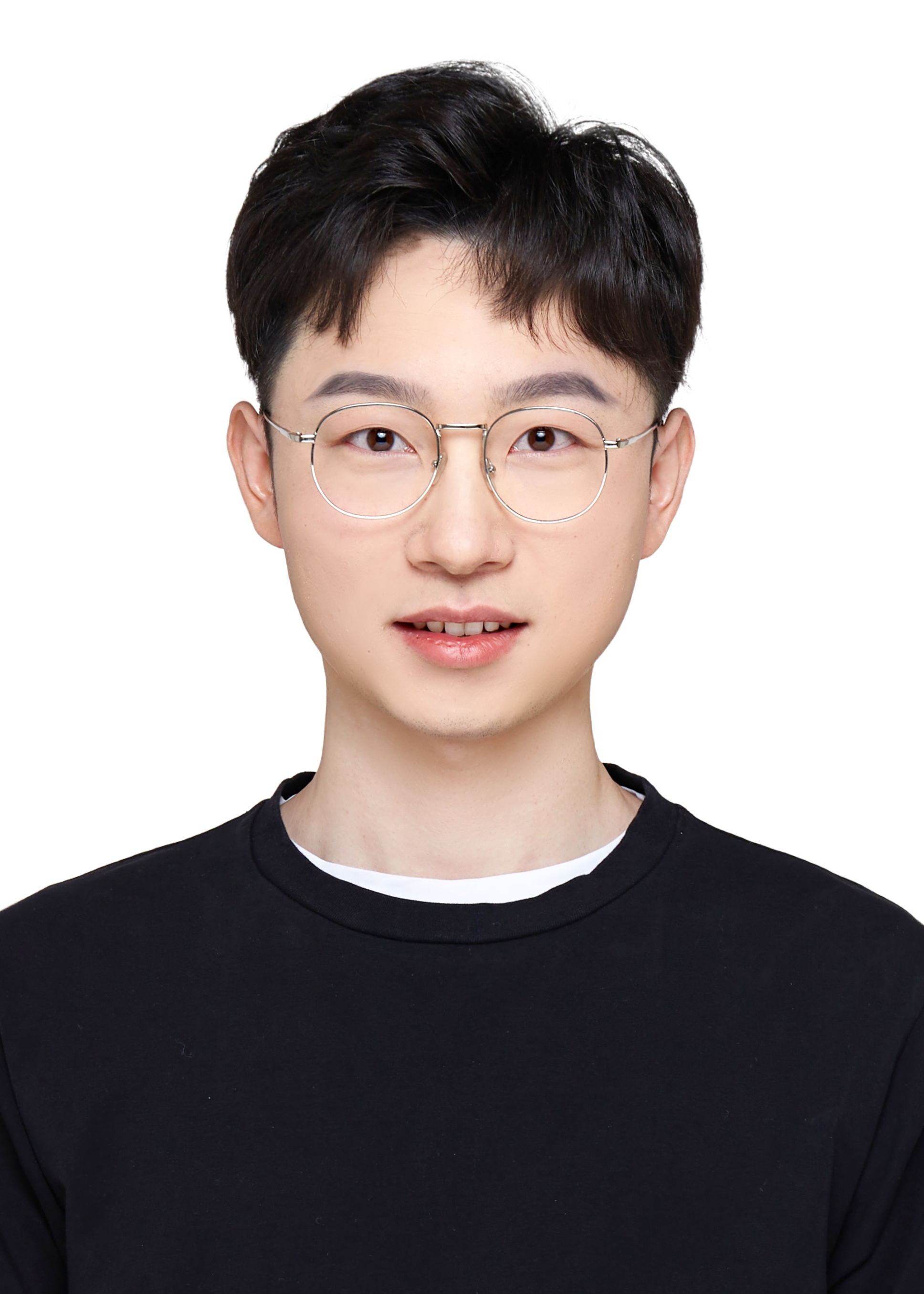}}]{Sun Long}
(Member, IEEE) received the B.S. degree in communication engineering from Xidian University, Xi’an, China, in 2018, and the Ph.D. degree from Xidian University, Xi’an, China, in 2024. 

He is currently a postdoctoral researcher of Key Laboratory of Intelligent Perception and Image Understanding of Ministry of Education, School of Artificial Intelligence, Xidian University, Xi'an, China. His research interests include computer vision and deep learning.
\end{IEEEbiography}


\begin{IEEEbiography}[{\includegraphics[width=1in,height=1.25in,clip,keepaspectratio]{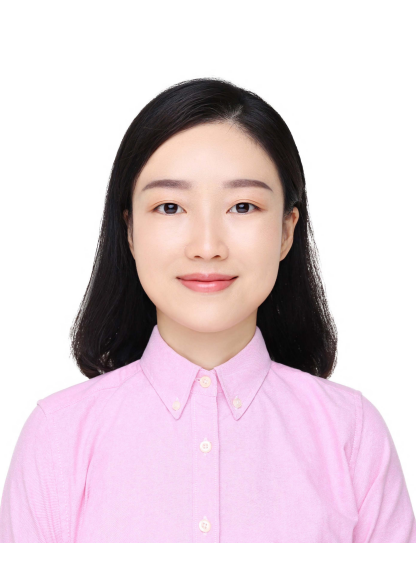}}]{Wenping Ma}
(Senior Member, IEEE) received the B.S. degree in computer science and technology and the Ph.D. degree in pattern recognition and intelligent systems from Xidian University, Xi’an, China, in 2003 and 2008, respectively. She is currently an Associate Professor with the School of Artificial Intelligence, Xidian University. Her research interests include natural computing and intelligent image processing. Dr. Ma is a member of Chinese Institute of Electronics (CIE).
\end{IEEEbiography}


\begin{IEEEbiography}[{\includegraphics[width=1in,height=1.25in,clip,keepaspectratio]{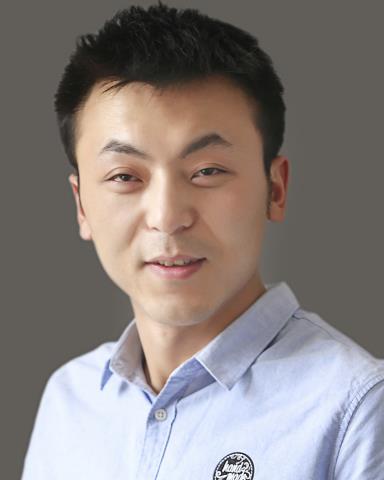}}]{Xu Liu}
(Senior Member, IEEE) received the B.S. degree from the North University of China, Taiyuan, China, in 2013, and the Ph.D. degree from Xidian University, Xi’an, China, in 2019. He is currently an Associate Professor of Huashan Elite with the Key Laboratory of Intelligent Perception and Image Understanding of Ministry of Education, School of Artificial Intelligence, Xidian University. 

His research interests include machine learning and image processing. 
\end{IEEEbiography}


\begin{IEEEbiography}[{\includegraphics[width=1in,height=1.25in,clip,keepaspectratio]{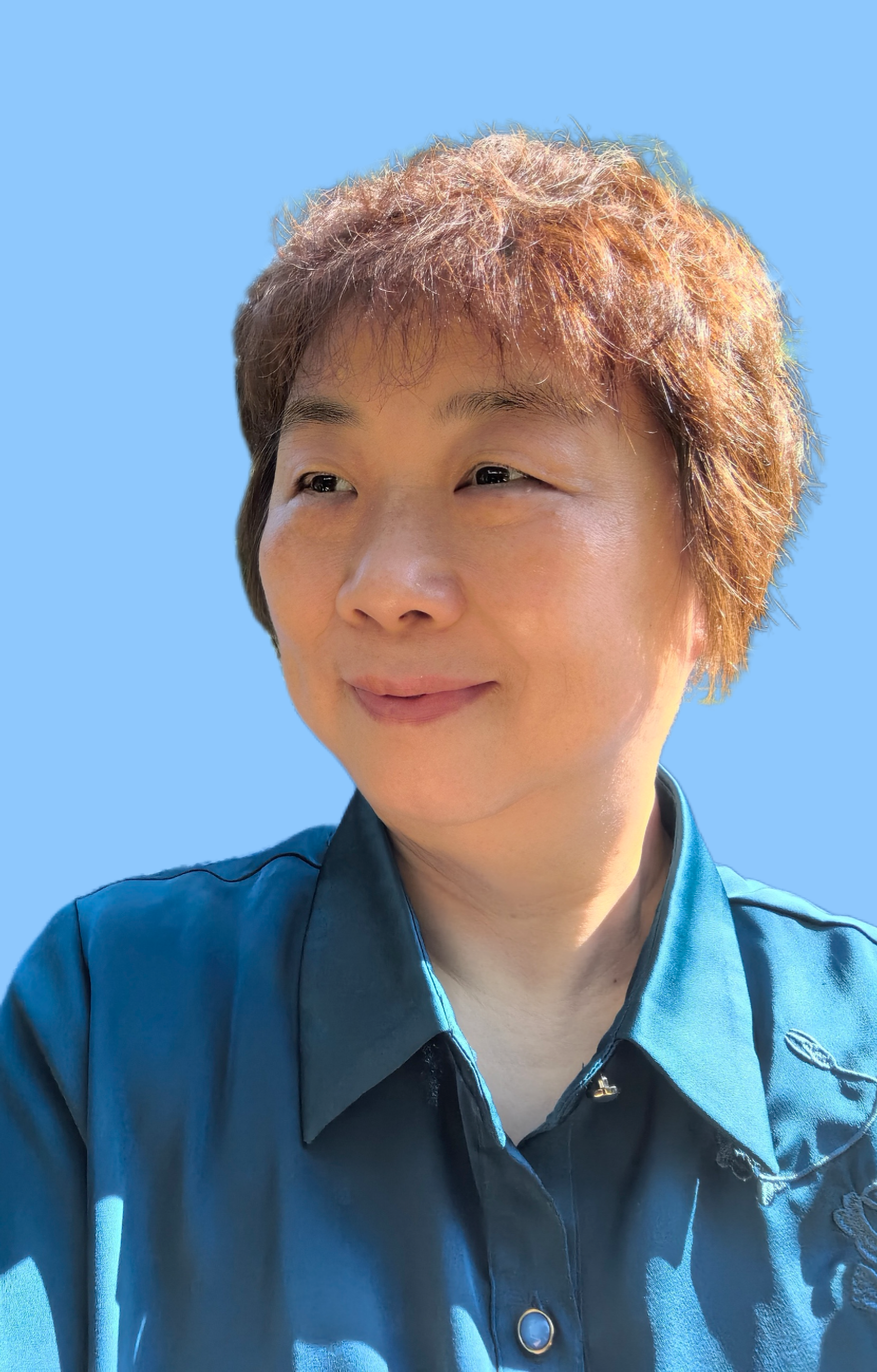}}]{Fang Liu}
(Senior Member, IEEE) received the B.S. degree in computer science and technology from Xi’an Jiaotong University, Xi’an, China, in 1984, and the M.S. degree in computer science and technology from Xidian University, Xi’an, in 1995. She is currently a Professor at Xidian University. She has authored or coauthored five books and over 80 papers. 

Her research interests include pattern recognition, machine learning, and evolutionary computation.
\end{IEEEbiography}

\end{document}